\documentclass[lettersize,journal]{IEEEtran}

\usepackage{amsmath,amssymb,amsfonts}
\usepackage{graphicx}
\usepackage{textcomp}
\usepackage{algorithm}
\usepackage[noend]{algpseudocode}
\usepackage{setspace}
\usepackage{enumitem}

\DeclareMathOperator{\argmax}{arg\,max}
\usepackage{tabularx}
\usepackage{multirow}
\usepackage{multicol}
\usepackage{diagbox}
\usepackage{adjustbox}

\usepackage{caption}
\usepackage{subcaption}
\usepackage{comment}
\usepackage{placeins}

\usepackage{xcolor}

\usepackage[numbers,sort&compress]{natbib}

\usepackage{array}
\newcolumntype{P}[1]{>{\centering\arraybackslash}p{#1}}
\newcommand{\quotes}[1]{``#1''}

\usepackage{subcaption}

\begin{document}

\title{Vision Transformer with Adversarial Indicator Token against Adversarial Attacks in Radio Signal Classifications }

\author{Lu Zhang, Sangarapillai Lambotharan,~\IEEEmembership{Senior Member, IEEE}, Gan Zheng,~\IEEEmembership{Fellow, IEEE}, Guisheng Liao,~\IEEEmembership{Senior Member, IEEE}, Xuekang Liu, Fabio Roli,~\IEEEmembership{Fellow, IEEE}, Carsten Maple
\thanks{\textit{(Corresponding author: Lu Zhang.)}\\ \indent{Lu Zhang is with School of Mathematics and Computer Science, Swansea university, Swansea, SA1 8EN, UK (e-mail: lu.zhang@swansea.ac.uk).}\\ \indent{Sangarapillai Lambotharan is with Institute for Digital Technologies, Loughborough University London, London, E20 3BS, UK (e-mail: s.lambotharan@lboro.ac.uk).}\\\indent{Gan Zheng is with School of Engineering, University of Warwick, Coventry, CV4 7AL, UK (e-mail: gan.zheng@warwick.ac.uk).}\\ \indent{Guisheng Liao is with School of Electronic Engineering, Xidian University, Xi’an, 710071, People’s Republic of China (e-mail: liaogs@xidian.edu.cn).}\\ \indent{Xuekang Liu is with the Department of Electrical and Electronic Engineering, Faculty of Engineering, Imperial College London, London, SW7 2AZ, U.K. (e-mail: xuekangliu@ieee.org).}\\ \indent{Fabio Roli is with Department of Informatics, Bioengineering Robotics, and Systems Engineering, University of Genova, Genoa, 16145, Italy (e-mail: fabio.roli@unige.it).}\\ \indent{Carsten Maple is with Warwick Manufacturing Group, University of Warwick, Coventry, CV4 7AL, UK (e-mail: cm@warwick.ac.uk).}}}

\maketitle
\begin{abstract}
The remarkable success of transformers across various fields such as natural language processing and computer vision has paved the way for their applications in automatic modulation classification, a critical component in the communication systems of Internet of Things (IoT) devices. However, it has been observed that transformer-based classification of radio signals is susceptible to subtle yet sophisticated adversarial attacks. To address this issue, we have developed a defensive strategy for transformer-based modulation classification systems to counter such adversarial attacks. In this paper, we propose a novel vision transformer (ViT) architecture by introducing a new concept known as adversarial indicator (AdvI) token to detect adversarial attacks. To the best of our knowledge, this is the first work to propose an AdvI token in ViT to defend against adversarial attacks. Integrating an adversarial training method with a detection mechanism using AdvI token, we combine a training time defense and running time defense in a unified neural network model, which reduces architectural complexity of the system compared to detecting adversarial perturbations using separate models. We investigate into the operational principles of our method by examining the attention mechanism. We show the proposed AdvI token acts as a crucial element within the ViT, influencing attention weights and thereby highlighting regions or features in the input data that are potentially suspicious or anomalous. Through experimental results, we demonstrate that our approach surpasses several competitive methods in handling white-box attack scenarios, including those utilizing the fast gradient method, projected gradient descent attacks and basic iterative method. 

\end{abstract}


\section{Introduction}

The rapid evolution of the Internet of Things (IoT) and mobile networks is driven by the ever-increasing demands for ultra-reliable performance, low-latency responsiveness, seamless connectivity, and enhanced mobility \cite{cui2021integrating, mu2021machine, zhang2020device, wang2019energy}. As it stands, over 50 billion devices are estimated to be wirelessly interconnected, providing sophisticated services and extensive environmental sensing capabilities. This surge in connected devices is intensifying the demand for already scarce wireless communication spectrum resources, necessitating more intelligent communication technologies to effectively manage this growth. Automatic modulation classification (AMC) emerges as a pivotal intelligent enabling technology in this context \cite{10049409, 9672088, 10460305, 10335761}. This technology facilitates the classification of modulation schemes without prior knowledge of the received signals or channel characteristics, enabling more efficient management of spectrum resources, self-recovery and maintaining communication reliability, thus supporting the continued growth and deployment of massive IoT devices even in the absence of a centralized network architecture and management.

The aim of AMC is to automatically identify the modulation type of a received signal, including BPSK, QPSK, 8PSK, QAM16, QAM64, CPFSK, GFSK, PAM4, WBFM, AM-SSB, and AM-DSB, etc. Historically, AMC relied on meticulously derived features, which were determined by specialists along with specific classification guidelines \cite{ghasemzadeh2019accuracy, abu2018automatic}. These traditional approaches are straightforward to execute in real-world scenarios, but their reliance on manually selected features and fixed classification criteria make it difficult to adapt to new types of modulation. More recently, the remarkable successes of deep learning have led many researchers to explore various deep neural network (DNN) models for AMC \cite{usman2020amc, o2018over, krzyston2020high, liao2021sequential, hao2024meta, 10049409, ramjee2019fast, hou2022multi, dong2022lightweight, hao2024smtc, tonchev2022automatic, fu2022automatic}. For instance, convolutional neural networks (CNNs) \cite{usman2020amc} have been utilized for this purpose. Subsequent developments included convolutional long short-term deep neural networks (CLDNN), long short-term memory neural networks (LSTMs), and deep residual networks (ResNets) to enhance classification accuracy \cite{ramjee2019fast}. Innovations also include a complex CNN \cite{hou2022multi} designed to identify signal spectrum information and a spatio-temporal hybrid deep neural network \cite{dong2022lightweight} that leverages multi-channel and multi-function blocks for AMC. A graph convolutional network (GCN) was proposed in \cite{tonchev2022automatic} which achieves comparable results to other existing approaches. Additionally, to minimize communication overhead, a new learning framework combining ensemble learning and decentralized learning \cite{fu2022automatic} was introduced. Moreover, inspired by the remarkable success of transformers in computer vision \cite{dosovitskiy2020image, wang2021pyramid, graham2021levit}, the authors of \cite{hamidi2021mcformer} has applied transformer models to AMC, achieving significant improvements in performance over previous methods. 

\begin{figure}[ht]
\centering
\includegraphics[width=\columnwidth]{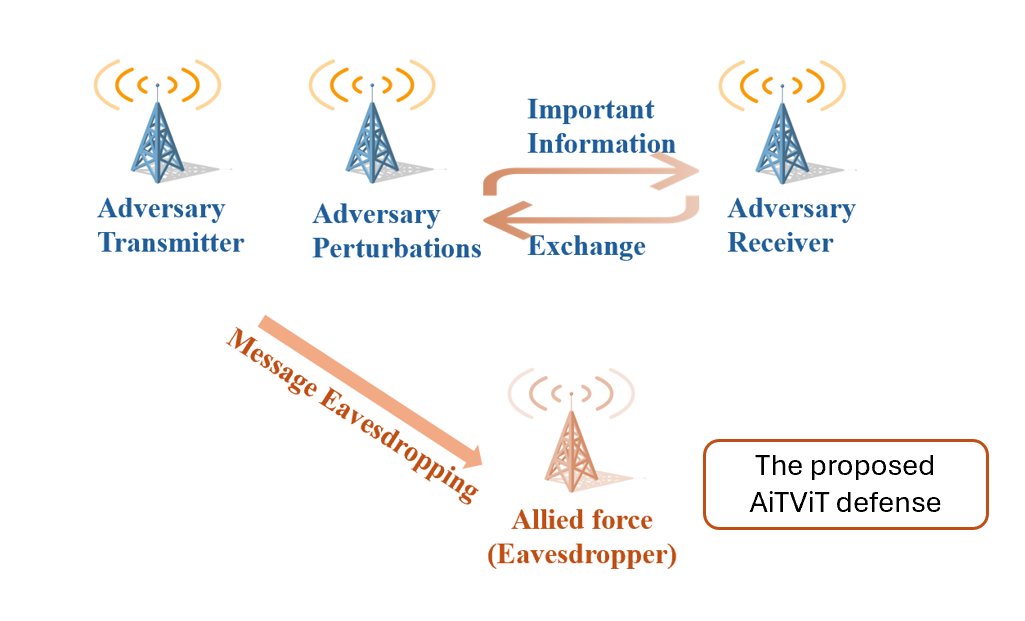}
  \caption{{A scenario of the adversarial attacks in modulation classification.}}
  \label{fig:eavesdropper}
\end{figure}

Despite the impressive performance of DNNs, numerous studies have highlighted their susceptibility to adversarial attacks. These are small, intentionally designed perturbations to the input data that can cause incorrect classifications \cite{xu2020adversarial}. Such adversarial attacks have been demonstrated to disrupt the functionality of various machine learning applications, including object detection \cite{zhang2020contextual}, natural language processing \cite{morris2020textattack}, face recognition \cite{xu2022adversarial}, and malware detection \cite{ling2023adversarial}, etc. Especially, the use of fast gradient methods (FGM) has been found to significantly impair classification accuracy in AMC \cite{sadeghi2018adversarial, zhang2021countermeasures}. A typical threat scenario is illustrated in Figure \ref{fig:eavesdropper}. Specifically, in a networked battlefield, radio signals are often used by adversary units (adversary transmitter and receiver as shown in Figure \ref{fig:eavesdropper}) to exchange critical battlefield sensing data. In this scenario, allied forces, acting as eavesdroppers, can utilize AMC to identify the modulation scheme and intercept the communication between adversary units. To prevent allied forces from successfully eavesdropping on their messages, the adversary can introduce adversarial perturbations to the communication signals, making it difficult for allied forces to accurately determine the modulation type. Hence, in this case, an AMC system that is robust against adversarial attacks should be applied by allied force as proposed in this paper. AMC differs considerably from conventional image classification tasks in both data characteristics and practical requirements. While image classification benefits from rich spatial and semantic content, AMC typically operates on raw in-phase and quadrature (IQ) samples that lack natural spatial semantics and correlations found in images. To address this challenge, we adopt an attention-based transformer architecture, which is effective at capturing long-range dependencies and abstract patterns, enabling the model to learn temporal and modulation-specific features from non-visual input. Furthermore, transformer-based defenses are well-suited for IoT applications due to their ability to model the complex and diverse signal patterns encountered in IoT communications. Another critical consideration in AMC is the need for real-time operation in resource-constrained environments, where computational complexity reduction is of paramount importance. Our proposed method integrates adversarial indicator token into the transformer framework to detect adversarial attacks, achieving both robustness and architectural simplicity suitable for practical AMC deployment. To be more specific, in this paper, we propose a novel ViT architecture by introducing an adversarial indicator (AdvI) token to defend against adversarial attacks (for notational simplicity, we refer to our proposed architecture as AiTViT). Without a defense mechanism like AiTViT, every misclassification directly results in decoding failure. However, with an AiTViT-based defense in place, the allied force aims to either correctly classify the modulation or detect adversarial examples. Upon detecting adversarial transmissions, i.e., adversarial examples, the allied force will be able to recognize the existence of adversarial transmissions, and the modulation classification will be deemed unreliable. Consequently, this enables the allied force to conserve computational resources by avoiding futile attempts at signal decoding or to switch to alternative modulation detection techniques such as those based on higher-order statistics. In summary, our key contributions in this work are:

\setlength{\textfloatsep}{0.6mm}

\begin{itemize}
  \item We propose a novel ViT architecture (AiTViT) introducing a new concept known as AdvI token to detect adversarial attacks. To the best of our knowledge, this is the first work to propose an AdvI token in ViT to defend against adversarial attacks.
  \item Combining AT training method and a detection mechanism using AdvI token, we integrate a training time defense and running time defense in a unified neural network model, which reduces architectural complexity of the system as compared to the two separate models.
  \item We examine the impact of our proposed AiTViT on the attention weights, focusing on how it highlights regions or features in the input data that may be considered anomalous
  \item Generating white-box FGM, projected gradient descent (PGD) and basic iterative method (BIM) attacks, we show the significant performance advantage of our proposed AiTViT compared to relevant exiting works, i.e., normally trained ViT, the AT trained ViT, NR and ATARD defense system.
\end{itemize} 

The remainder of the paper is arranged as follows: Section \uppercase\expandafter{\romannumeral2} presents the related works. The proposed methodology is illustrated in Section \uppercase\expandafter{\romannumeral3}, followed by the results and discussions in Section \uppercase\expandafter{\romannumeral4}. Finally conclusions are drawn in Section \uppercase\expandafter{\romannumeral5}.

\section{Related works}
Researchers have proposed various countermeasures to address adversarial attacks in AMC. For instance, Sahay et al. \cite{sahay2021deep} introduced a deep ensemble defense that integrates multiple deep learning architectures trained on both time and frequency domain representations of received signals. However, their approach is specifically designed for black-box scenarios, emphasizing the transferability of adversarial examples across signal domains, which is different from the white-box scenario as considered in this paper. The work in \cite{shtaiwi2022mixture} introduces a defense technique that filters out adversarial samples before they reach the modulation classifier. This approach leverages mixture GANs (MGAN), where a separate GAN is trained for each modulation scheme. However, this method significantly increases computational resource requirements due to the need to train an individual GAN for every modulation scheme, which is particularly unattractive for IoT devices due to their resource limitations. A neural rejection (NR) technique and its variants were proposed to detect adversarial attacks in \cite{zhang2021countermeasures} and \cite{9734753}. The NR system works by extracting last feature layer information from a CNN network and feeding them into a connected support vector machine system to detect adversarial attacks. Although NR improves the robustness against adversarial attacks compared to undefended DNN, integrating a CNN with an SVM increases the architectural complexity of the system. This can complicate both the development and maintenance processes, as two different model types must be managed and optimized to work effectively together. In addition, the data processing pipeline becomes longer when using two separate models. The output from the CNN needs to be processed and then fed into the SVM, which can introduce additional latency. However, as our proposed AiTViT also works based on the principle of anomaly detection which employs AdvI token to detect adversarial examples, we consider the NR system in \cite{zhang2021countermeasures} as one of the most competing works for performance comparison. 

Furthermore, in \cite{maroto2022safeamc} an adversarial training (AT) based defense was proposed in AMC area. As adversarial training (AT) is widely regarded as one of the most effective methods to defend against adversarial attacks \cite{zhu2021efficient, dolatabadi2022}, we include the AT-trained ViT as one of the most competitive works. Another highly relevant work is \cite{9922665}, where an attention-based adversarial robustness distillation method (ATARD) was proposed in AMC to transfer robustness to a compact transformer model by learning the adversarial attention map from a robust large transformer model. Since our work also involves a transformer-based defense, we consider the ATARD system in \cite{9922665} as one of the most competing works for performance comparison. ATARD system reduces the computational complexity of the model, however, it considers a training time defense only which shows lower robustness when adversarial perturbation is large. In this paper, to address the above issues, we propose a novel ViT architecture by introducing an AdvI token to detect adversarial attacks. In the meantime, this token acts as a crucial element within the ViT, influencing attention weights and thereby highlighting regions or features in the input data that are potentially suspicious or anomalous. To the best of our knowledge, this approach has not been explored in the literature, even in the broader context.

\begin{figure*}[ht]
\centering
\includegraphics[width=0.9\textwidth]{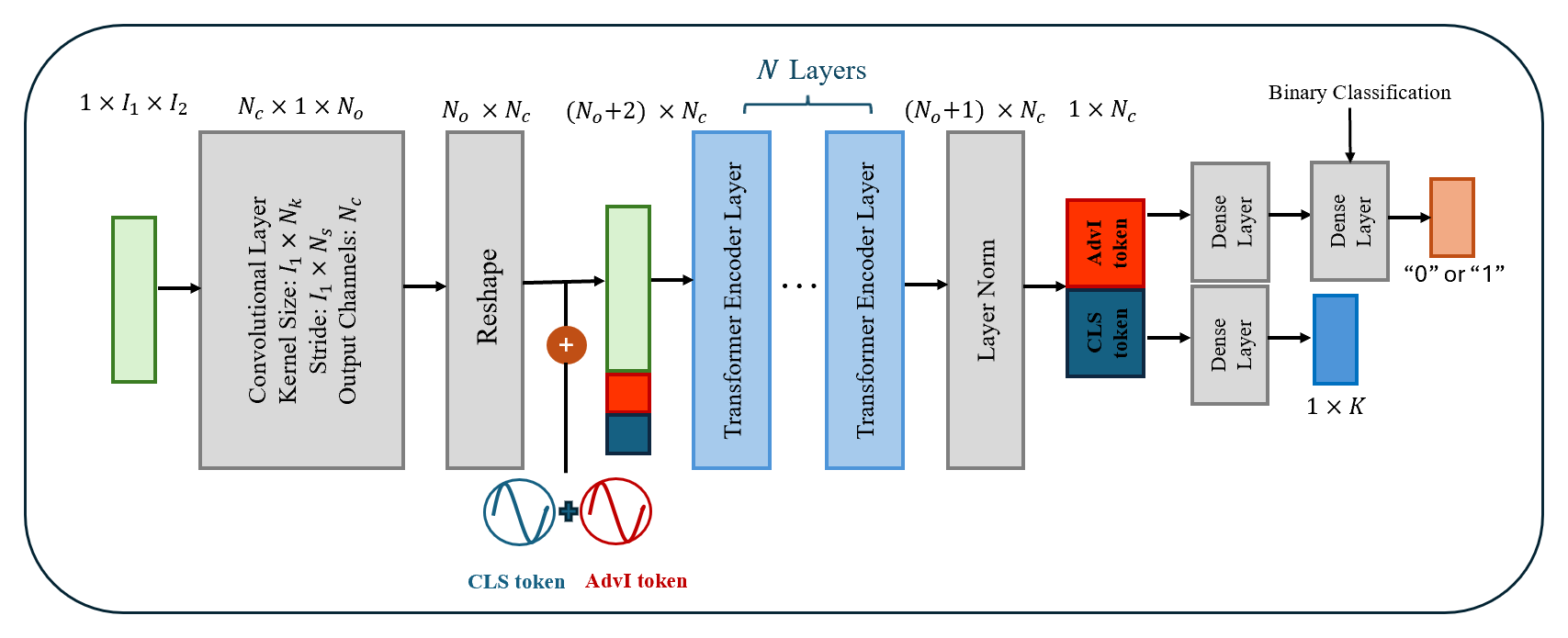}
  \caption{The architecture of the proposed AiTViT.}
  \label{fig:trannsformer}
\end{figure*}

\section{The proposed AiTViT}

Before providing specifics, we first introduce the ViT architecture used for AMC. The goal of ViT based AMC is to automatically identify the modulation type of a received signal, including BPSK, QPSK, 8PSK, QAM16, QAM64, CPFSK, GFSK, PAM4, WBFM, AM-SSB, and AM-DSB, etc. A ViT \cite{dosovitskiy2020image} is a neural network architecture originally designed for image recognition. In \cite{hamidi2021mcformer}, the authors introduced the ViT into AMC for the first time. Unlike traditional CNNs, ViTs divide a radio signal into small patches, treating these patches as a sequence. Each patch is converted into a vector, and a special "classification token" (CLS token) is included in the sequence to gather information from all patches for the final classification task. The CLS token is a learnable embedding that interacts with all other patches to capture a summary of the entire input. This sequence, including the CLS token, is processed through a Transformer encoder, and the output corresponding to the CLS token is used for modulation classification.

However, as demonstrated in \cite{lu2022Icassp}, the ViT-based AMC systems are vulnerable to adversarial attacks. To improve the adversarial robustness of ViT-based AMC systems, we propose a novel ViT architecture in this paper, referred to as AiTViT, by introducing an AdvI token to detect adversarial attacks. The motivation stems from the fact that ViTs leverage attention mechanisms to interpret and process input samples. However, as highlighted in \cite{fu2022patch}, the differences in attention maps generated by clean and adversarial inputs in the deeper layers of ViTs remain relatively small. By incorporating an AdvI token, we aim to enable the ViT to distinguish between the attention patterns of adversarial and benign samples during training. This allows the ViT to focus on the most critical regions of the input, thereby improving its capability to identify and respond to adversarial perturbations. To the best of our knowledge, this is the first work to propose an AdvI token into ViT for defending against adversarial attacks. In addition, a novel loss function which combines AT training technique and a detection mechnism, is proposed; In this way, a training time defense and running time defense are integrated in a unified neural network model, which reduces architectural complexity of the system as compared to the two separate models. We analyzed the effect of our proposed ViT on the attention weights, concentrating on its ability to emphasize areas or features in the input radio signal that could be identified as anomalous. Finally, to evaluate the performance of the proposed AiTViT defense system, white-box FGM, PGD and BIM attacks are designed, taking into account the unique architecture of the proposed AdvI token.

\subsection{The proposed architecture}
Now we provide details of our proposed AiTViT architecture. The key difference from ViT is the addition of the AdvI token, which complements the CLS token and is specifically designed to detect the presence of adversarial attacks. The AdvI token is a trainable embedding that interacts with all other patches, including the CLS token, by integrating information from the entire input. After passing through the transformer encoder layers, the output associated with the AdvI token is used to determine the presence of adversarial perturbations, with a binary classification indicating whether the input sample is benign (Class \quotes{0}) or contains adversarial perturbations (Class \quotes{1}). Additionally, during training, the AdvI token influences attention weights, helping to highlight regions or features in the input data that may be suspicious. This mechanism enables the proposed AiTViT to achieve enhanced robustness against adversarial perturbations.

The architecture for AiTViT is shown in Figure \ref{fig:trannsformer}. The input consists of I and Q components of the received communication signals combined to form a two-dimensional single-depth image ($1\times I_{1}\times I_{2}$). This is initially processed through a convolutional layer and subsequently reshaped, resulting in $N_{0}$ patch embeddings. The value of $N_{0}$ is determined by the formula $N_{0}=\frac{I_{2}-N_{k}}{N_{s}}+1$. Each patch embedding is of size $1\times N_{c}$, with $N_{c}$ set to 128 in this scenario. Two additional learnable embeddings, the CLS token and the AdvI token, are then added to the start of the patch embeddings array. The final state of the CLS token and the AdvI token at the end of the transformer encoder is used to represent the signal and indicate the existence of the adversarial attacks, respectively. The CLS token and the AdvI token share the same dimensionality of each patch embedding and their parameters are optimized during training through backpropagation. Altogether, the set of embeddings, including the CLS token and AdvI token, is represented as $\mathbf{z_{0}}$, comprising $N_{0}+2$ patch embeddings. These embeddings are then passed through $N$ transformer encoder layers (as $N=4$ here), which maintain the same dimensional structure as the input ($(N_{0}+2)\times N_{c}$). 

The encoder layer, a crucial component of the AiTViT network, is made up of two distinct sub-layers. The initial sub-layer employs a multi-head self-attention (MSA) mechanism, while the second is a position-wise fully connected feed-forward network. Each sub-layer is wrapped with a layer normalization, and this is coupled with a residual connection. Specifically, the output from the encoder layer is denoted as $\mathbf{z_{n}}$.
\begin{equation}
\label{equ:transformer encoder1}
\mathbf{z_{n}'}=\textup{SA}(\textup{LN}(\mathbf{z_{n-1}}))+\mathbf{z_{n-1}}, ~~~~~n=1,...N
\end{equation}
\begin{equation}
\label{equ:transformer encoder2}
~~~~\mathbf{z_{n}}=\textup{FFN}(\textup{LN}(\mathbf{z_{n-1}'}))+\mathbf{z_{n-1}'},~~~~~n=1,...N
\end{equation}

The conventional self-attention (SA) mechanism operates by transforming a query and associated key-value pairs into an output. As detailed in \eqref{equ:attention1}, SA produces this output by calculating the weighted sum of the values $\mathbf{V}$, with weights $\mathbf{A}$ assigned to each value. The weight $\mathbf{A}$, referred to as the attention, is derived from the function involving the query $\mathbf{Q}$ and the keys $\mathbf{K}$, each having dimension $d_{k}$. Particularly, $\mathbf{A}$ is computed through scaled dot products between the query and the keys, which is then processed through a softmax function, as depicted in \eqref{equ:attention2}.
\begin{equation}
    \label{equ:attention1}
     \textup{SA}(\mathbf{Q},\mathbf{K},\mathbf{V})= \mathbf{A}\mathbf{V}
\end{equation}
\begin{equation}
    \label{equ:attention2}
     \mathbf{A} = \textup{softmax}(\frac{\mathbf{Q}\mathbf{K}^{\textup{T}}}{\sqrt{d_{k}}})
\end{equation}
MSA extends SA by executing SA operations $h$ times simultaneously. The results from these operations are then merged and subsequently projected. Thus, the output of MSA is described as follows:
\begin{equation}
    \label{equ:msa}
     \textup{MSA}(\mathbf{Q},\mathbf{K},\mathbf{V})= [\textup{SA}_{1}; \textup{SA}_{2};...;\textup{SA}_{h}]\mathbf{U}_{\textup{MSA}},
\end{equation}
where $\mathbf{U}_\textup{{MSA}}$ is a linear projection matrix. Following these transformations, a layer normalization \cite{dosovitskiy2020image} process is applied, particularly to the CLS token and AdvI token output. Subsequently, the output of CLS token is sent through a dense layer, culminating in a $1\times K$ dimensional vector that stands for the likelihoods of $K$ different classes for the input signal. In the meantime, the output of AdvI token is fed to two dense layers, which performs a binary classification to detect adversarial attacks. The initial dense layer in the network produces an output of size $N_{i}$, which is set to 32, and it is coupled with a GeLU activation function for nonlinear processing. Following this, the second layer outputs $N_{b}$ dimensions, where $N_{b} = 2$. Through its output, we obtain two prediction scores for the input sample. If it belongs to Class "0", it means the sample is benign, otherwise, it indicates the presence of adversarial perturbations.

\subsection{The proposed training method}
In this section, we detail our training methodology for the proposed AiTViT. Before delving into details, we first illustrate an adversarial training (AT) \cite{madry2017towards} technique. AT is a robustness-enhancing technique where a neural network is trained using a mixture of clean and adversarially perturbed data. The process involves generating adversarial attacks during the training phase and incorporating them into the training dataset alongside the original examples. This approach helps the network to learn more robust features that are less susceptible to slight, often imperceptible, perturbations designed to mislead the model. Formally, let $\mathbf{x}$ be a clean sample from the training dataset with its corresponding label $y$. An adversarial attack $\mathbf{x}^{adv}$ is generated by applying a small perturbation $\mathbf{\delta}$ to $\mathbf{x}$, where $\mathbf{\delta}$ is often obtained by maximizing the loss function $L$ used to train the network, constrained by $\left \| \mathbf{\delta} \right \|\leq \varepsilon $, with $\varepsilon $ being a small upper bound characterizing the worst case perturbations. This ensures that the perturbed example $\mathbf{x}^{adv}$ remains perceptually similar to $\mathbf{x}$. The network is then trained on both the clean and the adversarial examples. The objective function for AT can be expressed as:
\begin{equation}
\label{equ:loss function of AT}
\underset{\theta}{\operatorname{min}} \mathbb{E}_{(\mathbf{x},y)\sim D}[\alpha L(f_{\theta }^{1}(\mathbf{x}),y)+(1-\alpha) L(f_{\theta }^{1}(\mathbf{x}^{adv}),y)],
\end{equation}
where $\theta$ denotes the parameters of the model, $D$ is the distribution of training data, $f_{\theta }^{1}$ represents the classification model, and $\alpha$ is a hyperparameter that balances the importance of clean and adversarial examples in the training process.

Now based on AT, we describe our proposed training method which considers the proposed AdvI token within ViT architecture. Most adversarial defense strategies tend to compromise normal accuracy (i.e., the accuracy for classifying benign original samples). To address this issue, in our proposed method, we pretrained models using the normal training (NT) approach. This choice leverages the strengths of pretrained models to enhance the generalization performance of our system. The training algorithm of the proposed AiTViT is shown in Algorithm 1. Specifically, during each training iteration, we generate adversarial version of the examples using projected gradient descent (PGD) algorithm as in line 5. The PGD attack is an iterative method used to craft adversarial examples by repeatedly adjusting a sample in the direction that maximizes the model’s prediction error. Each modification is followed by a projection step to ensure the perturbations remain within a predefined bound, maintaining the sample's similarity to the original. Considering the training signals span an extensive range of SNR from -20dB to 18dB, we propose an adaptive method to generate PGD attacks, where the size of the perturbations generated by PGD attacks increases incrementally as the SNR rises. After adversarial version of the samples are obtained, we train our proposed AiTViT. Specifically, our objective function (i.e., loss function) comprises two components, $loss_{1}$ and $loss_{2}$. The first component, $loss_{1}$, is designed to ensure robust modulation classification by aligning the predictions of both the adversarial samples and the original samples as closely as possible with the ground truth label $y$. The objective of $loss_{1}$ can be written as follows:
\begin{equation}
\label{equ:loss1}
loss_{1} = L(f_{\theta }^{1}(\mathbf{x}),y)+ L(f_{\theta }^{1}(\mathbf{x}^{adv}),y),
\end{equation}
where $f_{\theta }^{1}(\cdot)$ outputs the predictions of $K$ different classes for the input signal after processing the information extracted from the CLS token as in Figure \ref{fig:trannsformer}, and $L$ is the cross entropy (CE) loss in this work. The objective of the CE loss function is to measure the discrepancy between the predicted probability distribution and the true distribution, typically the ground truth labels. The goal is to refine the model’s predictions to closely match the true labels, effectively reducing the cross-entropy between these distributions. Formally, the cross-entropy loss function can be expressed using the following equation:
\begin{equation}
\label{equ:CE}
L=-\sum_{i}y_{i}\textup{log}(p_{i}),
\end{equation}
where $i$ means each possible class, $y_{i}$ is the true label, and $p_{i}$ is the predicted probability that the model assigns to the corresponding class. In addition, the second component of the loss function, $loss_{2}$, is formulated to differentiate between adversarial and benign samples, i.e., it performs binary classification based on the information extracted by the AdvI token. Specifically, after processing through two dense layers, adversarial detection is achieved by encouraging the predictions of adversarial samples and benign samples to align closely with class '1' (denoted as $y_{a}$) and class '0' ($y_{b}$), respectively. Formally, the objective of the $loss_{2}$ can be written as follows:
\begin{equation}
\label{equ:loss2}
loss_{2} = L(f_{\theta }^{2}(\mathbf{x}),y_{b})+ L(f_{\theta }^{2}(\mathbf{x}^{adv}),y_{a}),
\end{equation}
where $f_{\theta }^{2}(\cdot)$ outputs the predictions of two different classes for the input signal after processing the information extracted from the AdvI token as shown in Figure \ref{fig:trannsformer}. Overall, the total loss we used for training our proposed AiTViT is the linear combination of $loss_{1}$ and $loss_{2}$.
\begin{equation}
\label{equ:loss2}
loss = \beta loss_{1} + (1-\beta )loss_{2},
\end{equation}
where $\beta$ is a hyperparameter that balances the importance of two loss functions. In our model, we use two different types of classification tasks: $loss_{1}$ handles an 11-class classification, while $loss_{2}$ deals with a binary classification for detecting adversarial examples. To better focus on accurately identifying these adversarial perturbations, we assign a higher weighting factor to $loss_{2}$, i.e., in this work, we use $\beta = 0.1$.

\begin{algorithm}[ht!]
\caption{The Proposed Training Method}\label{alg:ATARD}
\hspace*{\algorithmicindent}\textbf{Input: }
\begin{itemize}[leftmargin=1.1cm]
\item learning rate $\eta$
\item number of epochs $N$
\item batch size $m$, number of batches $M$
\item distribution of the training data D
\end{itemize}
\hspace*{\algorithmicindent}\textbf{Output: }adversarially robust AiTViT model $G$

\begin{algorithmic}[1]
\vspace{2mm}
\State \textbf{for} $epoch=1,...,N$ \textbf{do}
\State ~~~\textbf{for} mini-batch=$1,...,M$ \textbf{do}
\State ~~~~~~Sample a mini-batch $\left \{ (\mathbf{x}_{i},y_{i}) \right \}_{i=1}^{m}$ from $D$
\State ~~~~~~\textbf{for} $i=1,...,m$ (in parallel) \textbf{do}
\State ~~~~~~~~Generate adversarial sample $\mathbf{x}_{i}^{adv}$ of $\mathbf{x}_{i}$ using \hspace*{13mm} adaptive PGD generation methods.
\State ~~~~~~\textbf{end for}
\State ~~~~~$\theta \leftarrow \theta -\eta \triangledown _{\theta }\{ \beta[L(f_{\theta }^{1}(\mathbf{x}),y)+ L(f_{\theta }^{1}(\mathbf{x}^{adv}),y)] + \hspace*{23mm}(1-\beta)[L(f_{\theta }^{2}(\mathbf{x}),y_{b})+ L(f_{\theta }^{2}(\mathbf{x}^{adv}),y_{a})]\}$
\State ~~~\textbf{end for}
\State \textbf{end for} 
\end{algorithmic}
\end{algorithm}

\subsection{Generation of white-box FGM, PGD and BIM attacks}

In this section, we generate white-box FGM, PGD and BIM attacks to assess the robustness of the proposed AiTViT. Adversarial attacks can be categorized into three types based on the extent of the attacker's knowledge: zero-knowledge black-box attacks, limited-knowledge gray-box attacks, and perfect-knowledge white-box attacks. In a black-box attack, the attacker has no information about the defense system's architecture, parameters, and data. In contrast, in a white-box setup the attacker is fully aware of the system, and a gray-box attack involves partial knowledge of the system. Our focus on white-box attacks in this work helps establish a benchmark for the worst-case scenario, assessing how such attacks could potentially degrade the performance of the defense mechanisms under severe conditions \cite{Biggio2018}. In computer vision applications, ensuring that adversarial perturbations remain imperceptible involves limiting their magnitude. Similarly, in the domain of radio signal classification, the concept of imperceptibility is governed by the perturbation-to-noise ratio (PNR) which refers to the ratio of the power of the adversarial perturbation relative to the power of the noise. Specifically, a perturbation is considered imperceptible if the PNR does not exceed 0 dB, indicating that the perturbation's magnitude is at or below the level of the noise, as outlined in \cite{sadeghi2018adversarial, sahay2021deep, aristodemou2022investigation}.

The procedure for the white-box FGM attack generation is outlined in Algorithm \ref{alg:FGM} which is adapted from \cite{sadeghi2018adversarial}. The main difference is that when designing the objective, we need to consider the detection mechanism of AdvI token. Specifically, the adversarial perturbation consists of two components. As shown in line 1 of the algorithm, we calculates the gradient of the cross-entropy loss between the predicted classification probabilities, denoted as $f^{1}(\mathbf{x}_{0})$, and the ground truth label, $y$. In addition, we calculate the gradient of the cross-entropy loss between the predicted detection likelihood $f^{2}(\mathbf{x}_{0})$, and $y_a$. This gradient is calculated with respect to the input, $\mathbf{x}_{0}$. Line 2 calculates the normalized perturbation $g_{norm}$ through dividing the gradient $g$ by its $l_{2}$ norm. After obtaining the normalized perturbation, it is scaled by the allowed perturbation size $\varepsilon$ and subsequently added to the original data sample in line 3. 

In particular, consider an input signal $\mathbf{x}_{0}$ with specified PNR and SNR values, where SNR represents the power of the signal relative to the noise. The allowable $l_{2}$-norm for the perturbation, $\varepsilon$, can be determined by the following equation:
\begin{equation}
\label{equ:perturbation norm}
\varepsilon =\sqrt{\frac{\textup{PNR} \cdot  \left \| \mathbf{x}_{0} \right \|_{2}^{2}}{\textup{SNR}+1}}
\end{equation}
This equation is derived as follows: The RML dataset \cite{OShea2016}, denoted as $\mathbf{x}$, comprises signal with signal power $S$ and noise with noise power $N$. The total power $P_{x}$ of $\mathbf{x}$, due to the independence of signal and noise, is the sum of $S$ and $N$, i.e., $P_{x} = S + N$. Consequently, $\frac{P_{x}}{N} = \frac{S + N}{N} = \textup{SNR} + 1$ can be established, leading to $N = \frac{P_{x}}{\textup{SNR} + 1}$. Given $\varepsilon$ as the $l_{2}$-norm of the adversarial perturbation, the power of the perturbation becomes $\frac{\varepsilon^{2}}{L}$, where $L$ is the number of elements in the perturbation vector. Hence, it can be derived that $\textup{PNR} = \frac{\varepsilon^{2}/L}{N} = \varepsilon^{2} \cdot \frac{\textup{SNR} + 1}{P_{x}L}$. Solving for $\varepsilon$, we find $\varepsilon =\sqrt{\frac{\textup{PNR}\cdot P_{x}L}{\textup{SNR}+1}}$. This relationship confirms \eqref{equ:perturbation norm} by substituting $P_{x}L$ with the sample estimate $\left \| \mathbf{x}_{0} \right \|_{2}^{2}$.

\begin{algorithm}[ht!]
\caption{FGM attack for the proposed AiTViT}\label{alg:FGM}
\hspace*{\algorithmicindent}\textbf{Input: }
\begin{itemize}[leftmargin=1.1cm]
\item input $\mathbf{x}_{0}$ and its true label $y$ 
\item the estimated classification likelihood $f^{1}(\cdot)$ for the input $\mathbf{x}_{0}$
\item the estimated detection likelihood $f^{2}(\cdot)$ for the input $\mathbf{x}_{0}$
\item the class that indicates the adversarial attack $y_{a}$ 
\item permitted $l_{2}$-norm for the perturbation $\varepsilon$, allowed PNR and SNR
\item cross entropy loss $l$
\end{itemize}
\hspace*{\algorithmicindent}\textbf{Output: }$\mathbf{x}^\prime$: the adversarial attacks.

\begin{algorithmic}[1]

\State $\mathbf{g} = \triangledown _{\mathbf{x}_{0}}l(f^{1}(\mathbf{x}_{0}), y)+ \triangledown _{\mathbf{x}_{0}}l(f^{2}(\mathbf{x}_{0}), y_{a})$
\State $\mathbf{g}_{norm} = \mathbf{g}/\left \| \mathbf{g} \right \|_{2}$
\State $\mathbf{x}^\prime =\mathbf{x}_{0}+\varepsilon \cdot \mathbf{g}_{norm}$
\State \Return{$\mathbf{x}^\prime$}

\end{algorithmic}
\end{algorithm}

The PGD attack is recognized for its effectiveness as it capitalizes on local, first-order information regarding the network \cite{madry2017towards}. The generation of PGD attack for the proposed AiTViT is outlined in Algorithm \ref{alg:PGD} which is adapted from \cite{madry2017towards}. There are two main differences. The first is that we need to take into consideration the detection mechanism of AdvI token when designing the gradients. In addition, in terms of the termination condition, instead of only forcing the modulation classification fail, we need to add another constraint make the generated PGD attack avoid the detection of the AdvI token. Specifically, line 4 of the algorithm outlines the initial step where the gradients need to be computed which consists of two components. The first is the cross-entropy loss between the model's predicted probabilities $f^{1}(\mathbf{x}_{0})$, and the actual classification label $y$. The second is the cross-entropy loss between the predicted detection likelihood $f^{2}(\mathbf{x}_{0})$, and the true detection label $y_b$. This gradient serves as the basis for a standard gradient update to derive the new sample $\mathbf{x}^{*}$ as in line 5, in which $\eta_{0}$ = 0.001. Following this, in line 6, a projection procedure is applied to $\mathbf{x}^{*}$.This procedure ensures that the size of the adversarial perturbation does not exceed the predefined threshold $\varepsilon$, as stipulated by the equation \eqref{equ:perturbation norm}. The projection operation itself is defined through the optimization process outlined below.

\begin{equation}
    \begin{aligned}
    \label{equ:projection1}
    &\min_{\mathbf{x}^\prime}\left \| \mathbf{x}^\prime - \mathbf{x}^{*} \right \|_{2}^{2},\\&
    s.t.~\left \| \mathbf{x}^\prime -\mathbf{x}_{0}\right \|_{2}\leq \varepsilon \\
    \end{aligned}
\end{equation}
where $\mathbf{x}^{*}$ represents the sample updated following the standard gradient step, while $\mathbf{x}_{0}$ refers to the initial input signal. The solution to \eqref{equ:projection1} is calculated as follows:
\begin{equation}
    \label{equ:projection11}
     \mathbf{x}^\prime = \mathbf{x}_{0}+\frac{\mathbf{x}^{*}-\mathbf{x}_{0}}{max(\varepsilon, \left \| \mathbf{x}^{*}-\mathbf{x}_{0} \right \|_{2})}\cdot \varepsilon.   
\end{equation}
Nevertheless, we utilize \eqref{equ:projector} as the projection method to ensure that the $l_{2}$ -norm of the adversarial perturbation produced $\left \| \mathbf{x}^\prime-\mathbf{x}_{0} \right \|_{2}$, equals $\varepsilon$. This approach facilitates the specification of a particular PNR for analyzing performance.

\begin{equation}
\label{equ:projector}
\mathbf{x}^\prime=\mathbf{x}_{0}+\frac{\varepsilon \cdot (\mathbf{x}^{*}-\mathbf{x}_{0})}{\left \| \mathbf{x}^{*}-\mathbf{x}_{0} \right \|_{2}}
\end{equation}
Ultimately, the loop stops  once the criterion in line 7 is satisfied, specifically when the predicted classification label of the generated adversarial sample does not match the actual label, and the predicted detection label is not equal to the adversarial indication class.

\begin{algorithm}[ht!]
\caption{PGD attack for the proposed AiTViT}\label{alg:PGD}
\hspace*{\algorithmicindent}\textbf{Input: }
\begin{itemize}[leftmargin=1.1cm]
\item the number of classes $K$
\item input $\mathbf{x}_{0}$ and its true label $y$ 
\item the step size $\eta_{0}$
\item the estimated classification likelihood $f^{1}(\cdot)$ for the input $\mathbf{x}_{0}$
\item the estimated detection likelihood $f^{2}(\cdot)$ for the input $\mathbf{x}_{0}$
\item the class that indicates the adversarial attack $y_{a}$ 
\item permitted $l_{2}$-norm for the perturbation $\varepsilon$, allowed PNR and SNR
\item cross entropy loss $l$
\item a projector $\Pi$ on the $l_{2}$-norm constraint $|| \mathbf{x}^\prime-\mathbf{x}_{0}||_{2}\leq \varepsilon$, where $\varepsilon =\sqrt{\frac{\textup{PNR} \cdot  \left \| \mathbf{x}_{0} \right \|_{2}^{2}}{\textup{SNR}+1}}$
\end{itemize}
\hspace*{\algorithmicindent}\textbf{Output: }$\mathbf{x}^\prime$: the adversarial attacks.

\begin{algorithmic}[1]

\State $\mathbf{x}^\prime\gets \mathbf{x}_{0}$

\Repeat
\State $\mathbf{x}\gets \mathbf{x}^\prime$
\State $\mathbf{g} = \triangledown _{\mathbf{x}_{0}}l(f^{1}(\mathbf{x}_{0}), y)+ \triangledown _{\mathbf{x}_{0}}l(f^{2}(\mathbf{x}_{0}), y_{a})$
\State $\mathbf{x}^{*}\gets \mathbf{x} + \eta_{0} \cdot g$
\State $\mathbf{x}^\prime\gets \Pi (\mathbf{x}^{*})$

\Until $\argmax_{i}f^{1}_{i}(\mathbf{x}^\prime)\neq y$ \textbf{and} $\argmax_{i}f^{2}_{i}(\mathbf{x}^\prime) \neq y_{a}$

\State \Return{$\mathbf{x}^\prime$}

\end{algorithmic}
\end{algorithm}


The BIM attack \cite{kurakin2018adversarial} is a notable contribution that introduced physical world attacks. The generation of BIM attack for the proposed AiTViT is outlined in Algorithm \ref{alg:BIM} which is adapted from \cite{kurakin2018adversarial}. There are two key differences. First, when designing the gradients, it is necessary to account for the detection mechanism of the AdvI token. Additionally, for the termination condition, rather than solely aiming to cause modulation classification failure, an additional constraint must be incorporated to ensure that the generated BIM attack evades detection by the AdvI token. Specifically, line 4 describes the initial step, where gradients are calculated based on two components. The first component is the cross-entropy loss between the model's predicted probabilities, $f^{1}(\mathbf{x}_{0})$,  and the true classification label, $y$. The second component is the cross-entropy loss between the predicted detection likelihood, $f^{2}(\mathbf{x}_{0})$, and the true detection label $y_b$. These gradients are processed through a sign function and used in a gradient update to generate the new adversarial sample, $\mathbf{x}^{*}$, as outlined in line 5, where $\eta_{0}$ = 0.001. In line 6, a clipping procedure is applied to $\mathbf{x}^{*}$ to ensure that the $l_{\infty }$-norm of the adversarial perturbation does not exceed the predefined threshold $\tau$. The parameter $\tau$ is used because the $L_{\infty}$-norm of the perturbation size is considered in BIM attacks, which cannot be directly correlated with PNR values. The loop terminates when the condition in line 7 is met: specifically, when the predicted classification label of the adversarial sample differs from the true label, and the predicted detection label does not match the adversarial indication class.

\begin{algorithm}[ht!]
\caption{BIM attack for the proposed AiTViT}\label{alg:BIM}
\hspace*{\algorithmicindent}\textbf{Input: }
\begin{itemize}[leftmargin=1.1cm]
\item the number of classes $K$
\item input $\mathbf{x}_{0}$ and its true label $y$ 
\item the step size $\eta_{0}$
\item the estimated classification likelihood $f^{1}(\cdot)$ for the input $\mathbf{x}_{0}$
\item the estimated detection likelihood $f^{2}(\cdot)$ for the input $\mathbf{x}_{0}$
\item the class that indicates the adversarial attack $y_{a}$ 
\item permitted $l_{\infty }$-norm for the perturbation $\tau$
\item cross entropy loss $l$
\end{itemize}
\hspace*{\algorithmicindent}\textbf{Output: }$\mathbf{x}^\prime$: the adversarial attacks.

\begin{algorithmic}[1]
\State $\mathbf{x}^\prime\gets \mathbf{x}_{0}$
\Repeat
\State $\mathbf{x}\gets \mathbf{x}^\prime$
\State $\mathbf{g} = \triangledown _{\mathbf{x}_{0}}l(f^{1}(\mathbf{x}_{0}), y)+ \triangledown _{\mathbf{x}_{0}}l(f^{2}(\mathbf{x}_{0}), y_{a})$
\State $\mathbf{x}^{*}\gets \mathbf{x} + \eta_{0} \cdot sign(g)$
\State $\mathbf{x}^\prime\gets clip(x^{*}-x_{0}, -\tau, \tau)$
\Until $\argmax_{i}f^{1}_{i}(\mathbf{x}^\prime)\neq y$ \textbf{and} $\argmax_{i}f^{2}_{i}(\mathbf{x}^\prime) \neq y_{a}$
\State \Return{$\mathbf{x}^\prime$}

\end{algorithmic}
\end{algorithm}

\section{Results and Discussion}
In this section, we discuss and examine the results of our experiments. All algorithms were developed using PyTorch and run on an NVIDIA GeForce RTX 4080 GPU.

\subsection{Experimental setup}

In this work, we utilize the GNU radio ML dataset RML2016.19a \cite{OShea2016} and RDL2021.12 \cite{luan2022automatic}. The GNU radio ML dataset RML2016.10a comprises 220,000 entries, each representing a distinct modulation type at a specific SNR. It encompasses 11 different modulation types including BPSK, QPSK, 8PSK, QAM16, QAM64, CPFSK, GFSK, PAM4, WBFM, AM-SSB, and AM-DSB. These entries are spread across 20 different SNR levels, varying from -20 dB to 18 dB in increments of 2 dB. Each entry is made up of 256 dimensions, split between 128 in-phase and 128 quadrature components. The dataset is divided evenly for training and testing purposes. Compared to the RML dataset, the RDL dataset includes two types of noise: Gaussian noise and Alpha-stable distributed noise. Additionally, the RDL dataset considers Rician fading channel. It contains 110,000 samples generated for 11 different modulation types at ${\textup{SNR~= 10 dB}}$, with 90$\%$ of the samples used for training and the remaining 10$\%$ for testing. The inclusion of Alpha-stable noise introduces impulsive noise into the RDL dataset. To mitigate the impact of this noise, the received signals were pre-processed before being fed into the classifiers during both the training and testing phases. Specifically, a five-sample moving median filter was applied to the signal, and the standard deviation, $\sigma_{x}$, was computed. Any samples falling outside the range of $\pm \sigma_{x}$ were discarded. Finally, the data samples were normalized to prepare them for model training and evaluation.

For evaluating the robustness of the proposed AiTViT network to adversarial perturbations, we produce adversarial version of the samples using 1000 test samples at an SNR of 10 dB and we consider all the perturbed samples as adversarial attacks. We present two evaluation metrics to assess the performance of the proposed methods: accuracy and false negative rate (FNR). Accuracy measures the proportion of True Positives (TP) and True Negatives (TN) out of the total number of samples, as shown in Equation \eqref{equ:accuracy}:
\begin{equation}
    \label{equ:accuracy}
    \textup{accuracy} = \frac{\textup{TP + TN}}{total~samples},
\end{equation}
where total number of the samples includes TP, TN, False Positives (FP), and False Negatives (FN). We have 1000 benign samples and 1000 generated adversarial attacks, resulting in a total of 2000 samples. Accuracy measures the proportion of samples that are either correctly classified or correctly recognized as anomalous by the detection system. The definition of FNR is shown in Equation \eqref{equ:FNR} which measures the proportion of adversarial attacks that the classifier fails to detect: 
\begin{equation}
    \label{equ:FNR}
    \textup{FNR} = \frac{\textup{FN}}{\textup{TP+FN}}.
\end{equation}

We use the normal training (NT) trained ViT as the baseline and multiple defense systems in AMC from the literature as comparison, including the AT trained ViT, the NR system, and the ATARD defense. The details of these comparison defenses are provided and discussed in Section II.

\subsection{Attention visualization}
In this section, to understand how the proposed adversarial indicator token influences the attention weights, we present a visualization of the attention mechanisms within our proposed AiTViT to analyze how it handles adversarial attacks. Our focus is specifically on examining the attention weights between the CLS token with AdvI token and other signal patches. Additionally, as a baseline for comparison, we also provide a visualization of the attention mechanisms within conventional ViT architecture trained by the AT technique. For this analysis, we generate PGD attacks for our proposed AiTViT, with detailed generation algorithms provided in Section D. We also produce adversarial versions of the same original samples based on AT trained conventional ViT. During generation, we introduce substantial adversarial perturbations with a PNR of 0dB when SNR = 10dB. The goal is to analyze if our proposed AiTViT can concentrate its attention on the most useful features of adversarial attacks. To directly visualize the main features extracted by the network on the signal waveform, after extracting the attention weights of the network, we map it back onto the input adversarial sample to analyze the distribution of attention weights. Specifically, we first extract the attention weights of the network, particularly the attention weights between the CLS token with the AdvI token and other signal patches. To map this attention features to the input adversarial signal, we up-sample the extracted attention features to match the size of the input signal. We then reshape this matrix to the dimensions of the original I/Q data. This approach ensures that each raw I/Q data point is associated with a unique neural unit activation weight. We specifically prioritize the weights that are relatively larger, indicating areas where the model focused most intensively during the adversarial processing. To facilitate this analysis, we normalize the weights to scale between 0 and 1. We then disregard any weights below 0.5, setting them to zero, to maintain our focus exclusively on the higher-impact features as identified by the attention mechanism.

\begin{figure}[ht!]
\centering
\includegraphics[width=0.35\textwidth]{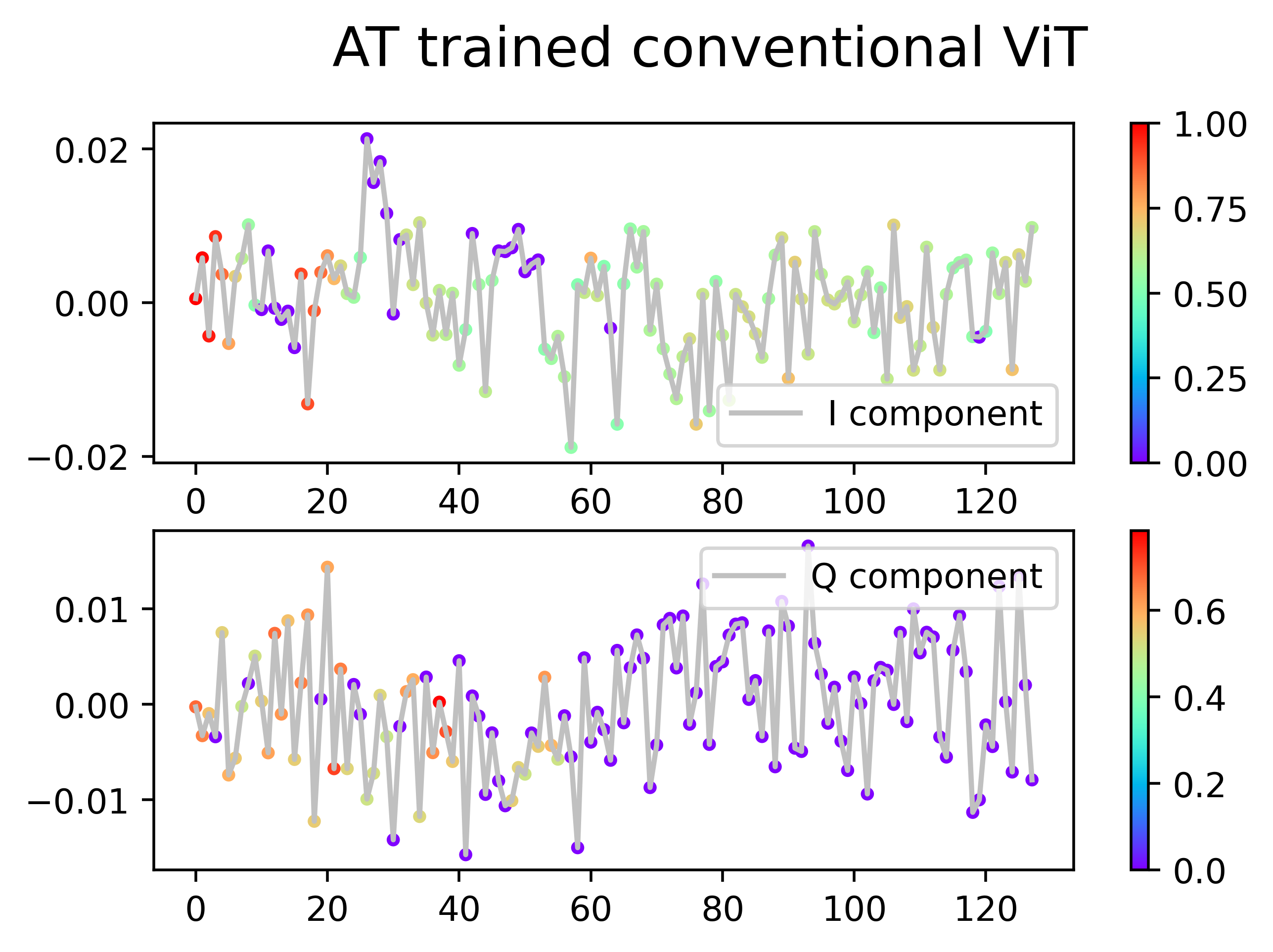}
\caption{An example of attention visualization for AM-SSB modulated signal within conventional ViT architecture.}
\label{fig:13_vit}
\end{figure}

\begin{figure}[ht!]
\centering
\includegraphics[width=0.35\textwidth]{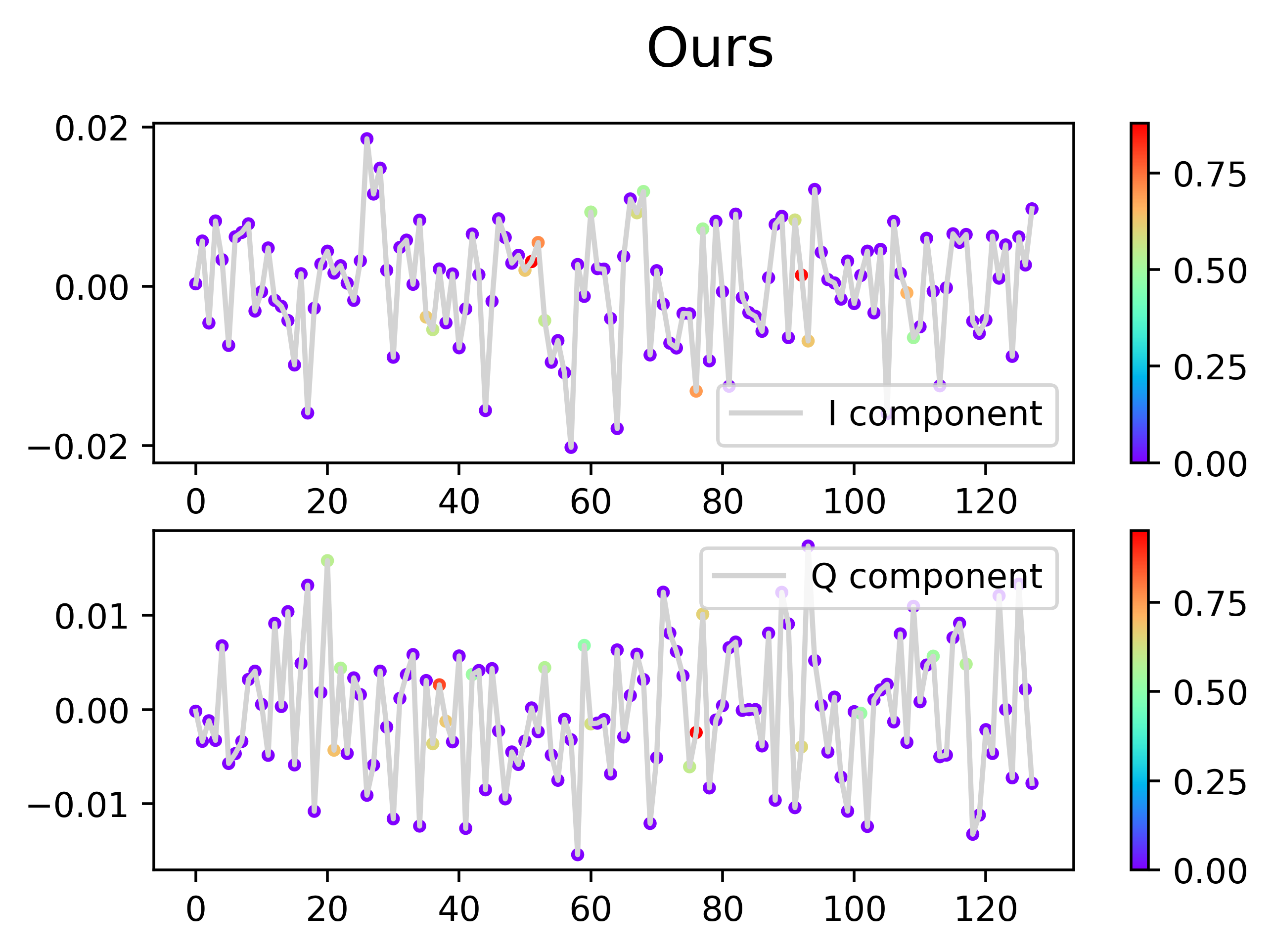}
\caption{An example of attention visualization for AM-SSB modulated signal within our proposed AiTViT.}
\label{fig:13_ours}
\end{figure}

\begin{figure}[ht!]
\centering
\includegraphics[width=0.35\textwidth]{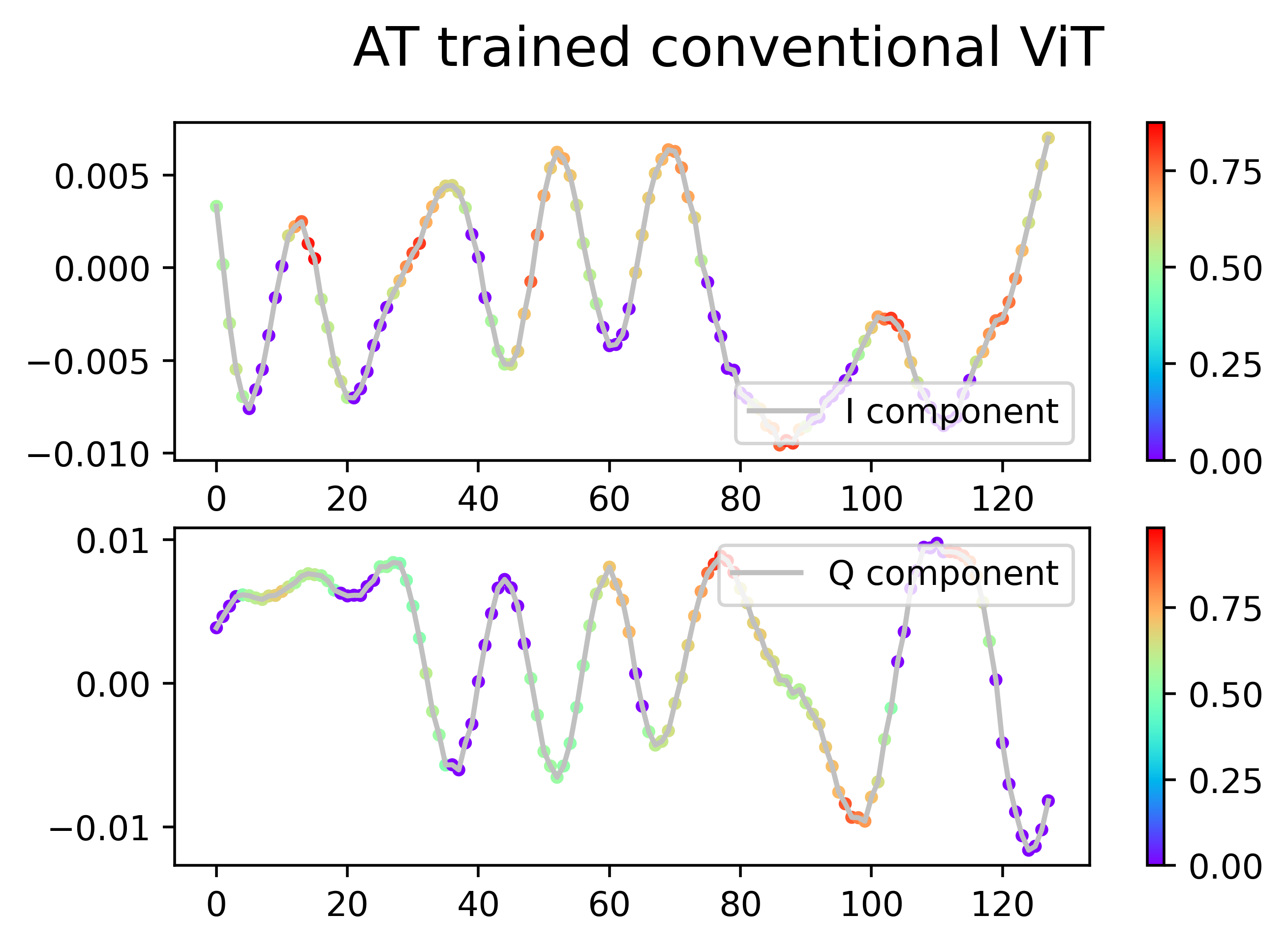}
\caption{An example of attention visualization for QAM16 modulated signal within conventional ViT architecture.}
\label{fig:30_vit}
\end{figure}

\begin{figure}[ht!]
\centering
\includegraphics[width=0.35\textwidth]{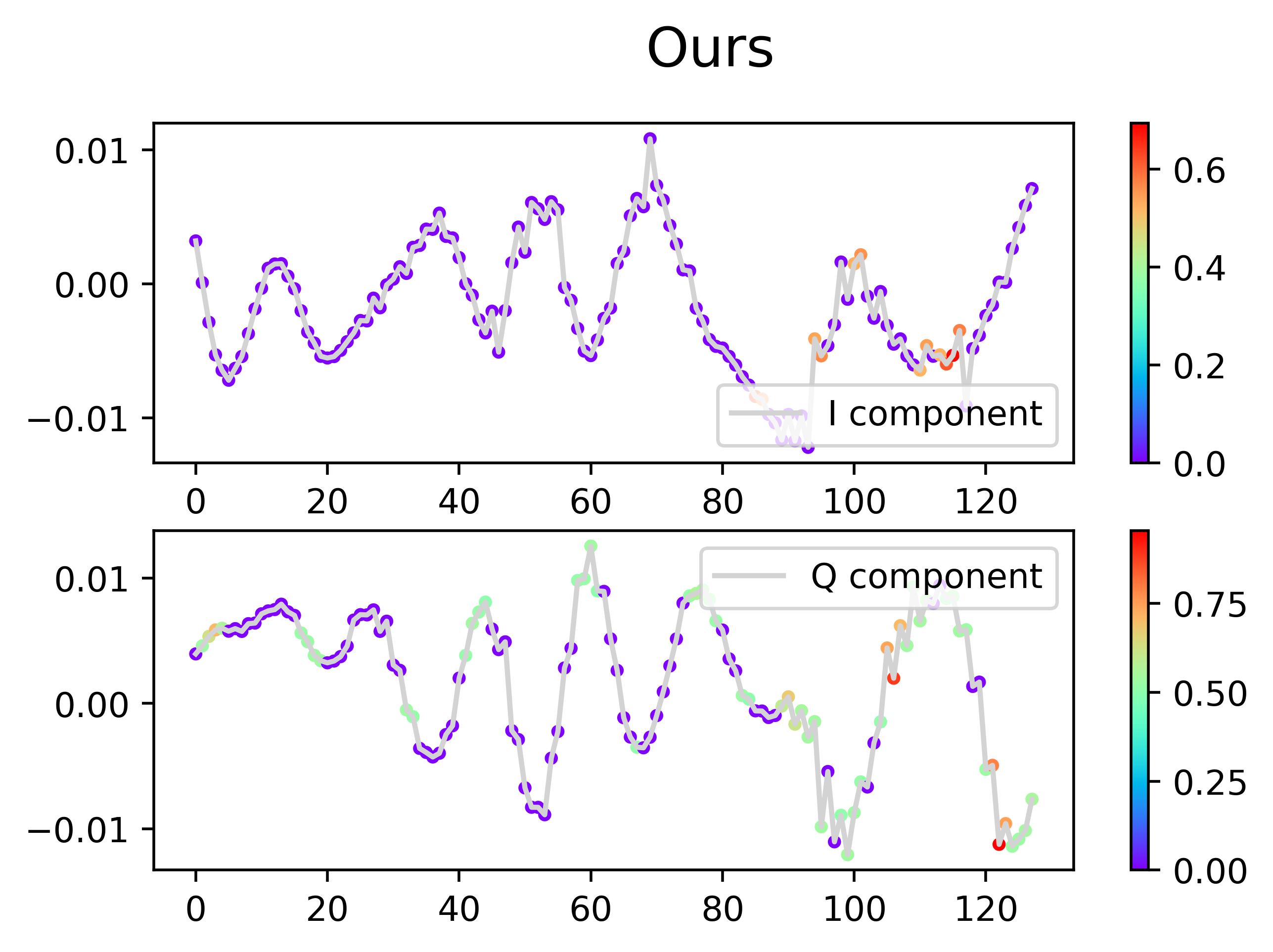}
\caption{An example of attention visualization for QAM16 modulated signal within our proposed AiTViT.}
\label{fig:30_ours}
\end{figure}

\begin{figure}[ht!]
\centering
\includegraphics[width=0.35\textwidth]{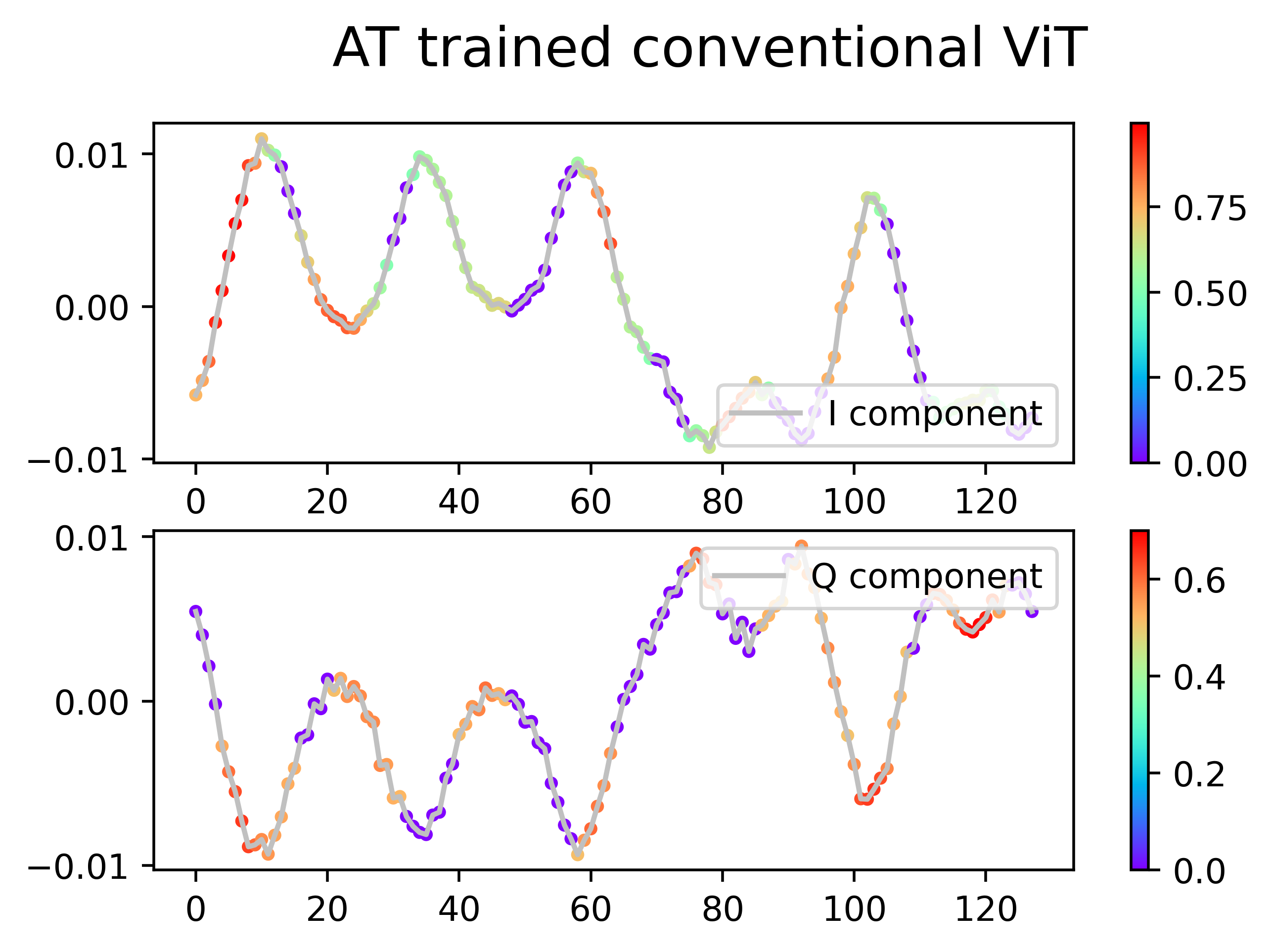}
\caption{An example of attention visualization for BPSK modulated signal within conventional ViT architecture.}
\label{fig:31_vit}
\end{figure}

\begin{figure}[ht!]
\centering
\includegraphics[width=0.35\textwidth]{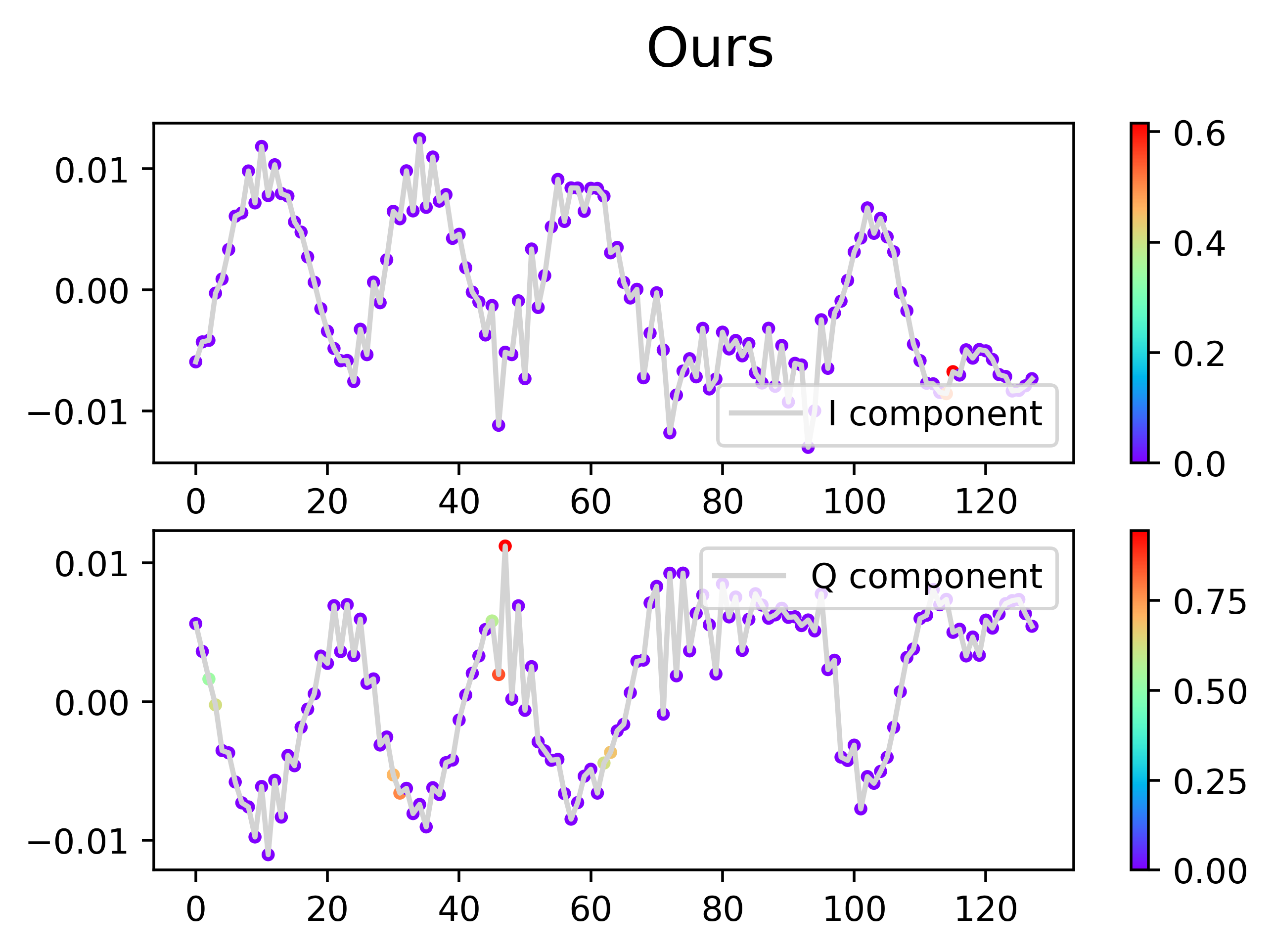}
\caption{An example of attention visualization for BPSK modulated signal within our proposed AiTViT.}
\label{fig:31_ours}
\end{figure}

We provide the visualization in Figure \ref{fig:13_vit} - Figure \ref{fig:31_ours}. The colorbar on the right side of the figures represents the value of the attention weights. It can be observed that the attention in a AT trained conventional ViT tends to be more dispersed across the entire input adversarial signals. In contrast, the attention weights of our proposed AiTViT are concentrated on fewer components of the signal, exhibiting a more concentrated attention distribution. This focused attention allows our model to better pinpoint and emphasize the critical features of the input sample, such as the phase, amplitude, and frequency of the signal, which are essential for accurate analysis. To further substantiate this finding, we compute the average number of attention elements with high attention weights (i.e., weights are not less than 0.5) across 1000 PGD attacks. The results showed that the average number for the AT trained conventional ViT was 165.48, while for our proposed AiTViT, it was significantly lower at 38.21. Clearly, our AiTViT demonstrates a more focused attention area when compared to the baseline, underscoring its ability to concentrate on the most relevant aspects of the input.

\subsection{Robustness results against white-box FGM, PGD and BIM attacks}
In this section, we provide simulation results to demonstrate the robustness of our proposed method against white-box FGM, PGD and BIM attacks. Considering a detection mechanism, our proposed AiTViT defends adversarial attacks by either correctly classifying them or detecting them. From Figure 9-12, it is clear that our proposed method can achieve higher robustness against FGM and PGD attacks for a range of PNR values. Specifically, as shown in Figure 9 and Figure 10, our proposed AiTViT can achieve significantly higher accuracy and lower FNR as compared to other 4 baseline defenses when PNR is greater than -15 dB. In particular, in Figure 9, when PNR = -10 dB, our proposed AiTViT achieves approximately 10$\%$ higher accuracy compared to the AT trained ViT and ATARD. Furthermore, it demonstrates a 26$\%$ improvement in accuracy compared to the NR system. In Figure 10, when PNR = 0 dB, our proposed AiTViT can achieve a very low FNR of 32$\%$ while the FNR of other four baseline methods increase to almost 100$\%$. The robustness of our method within the PNR range of -15 dB to 0 dB is primarily derived from the high detection rates. For instance, at a PNR of 0 dB, the detection rate can achieve 68$\%$, whereas at PNR = -10 dB, the rate increases to 80$\%$. 


In terms of FGM attacks, it can be seen from Figure 11 and Figure 12 that our proposed AiTViT outperforms the other four baseline methods. Specifically, the robustness is significantly higher when PNR is larger than -15 dB.  Notably, in Figure 11, when PNR = -10 dB, our proposed AiTViT achieves around 13$\%$ and 30$\%$ higher accuracy as compared to the AT trained ViT and NR, respectively. In addition, in Figure 12, when PNR = 0 dB, the FNR of our proposed AiTViT is nearly zero, whereas the FNR of other four baseline is significantly higher, ranging from 58$\%$ to 73$\%$. The robustness against FGM attacks for PNR of -10 dB and 0 dB mainly comes from the high detection performance our proposed method achieved. For example, our proposed AiTViT can have 85$\%$ and 89$\%$ detection rate when PNR = -10 dB and PNR = 0 dB respectively. Similar trend can be found for BIM attacks in Figure 13 and Figure 14. It can be seen that our proposed AiTViT outperforms the other four baseline methods for a range of adversarial perturbations size $\tau$. The parameter $\tau$ is used because the $L_{\infty}$-norm of the perturbation size is considered in BIM attacks, which cannot be directly correlated with PNR values. In particular, in Figure 13, when $\tau$ = 0.0012, our proposed AiTViT achieves approximately 23$\%$ higher accuracy compared to the AT trained ViT and ATARD. Furthermore, it demonstrates a 35$\%$ improvement in accuracy compared to the NR system. In Figure 14, our proposed AiTViT can achieve a very low FNR of ranging from 10$\%$ to 20$\%$, while the FNR of other four baseline methods increase to more than 90$\%$. 

For the RDL dataset, the accuracy and FNR against PGD attacks across a wide range of perturbation-to-generalized-noise ratio (PGNR) values for Rician fading channels is presented in Figure \ref{fig:PGD_10dB_RDL} and Figure \ref{fig:PGD_10dB_RDL_FNR}. Following the approach in \cite{luan2022automatic}, we assume identical power for Gaussian noise and alpha-stable noise. PGNR is defined as the ratio of the perturbation power to the alpha-stable noise power. Consequently, the ratio of perturbation power to Gaussian noise power is equivalent to PGNR, however, the ratio of perturbation power to the combined power of Gaussian and alpha-stable noise is given by $\textup{PGNR}/2$ (i.e., 3 dB less than PGNR in dB). Additionally, for a given SNR and PGNR value, the perturbation-to-signal ratio (PSR) can be calculated as $\textup{PSR} = \textup{PGNR} / \textup{SNR}$. It can be seen from Figure \ref{fig:PGD_10dB_RDL} and Figure \ref{fig:PGD_10dB_RDL_FNR} that our proposed AiTViT outperforms the NT and AT trained ViT. Notably, in Figure \ref{fig:PGD_10dB_RDL}, when PGNR = -10 dB, our proposed AiTViT achieves around 25$\%$ higher accuracy as compared to the AT trained ViT and NR.  In Figure \ref{fig:PGD_10dB_RDL_FNR}, our proposed AiTViT can achieve a very low FNR of 13$\%$, while the FNR of other four baseline methods increase to almost 100$\%$. To assess the generalization capabilities of our proposed method, we trained the AiTViT model using varying proportions of the total training dataset, ranging from 10$\%$ to 50$\%$. As depicted in Figure 17 and Figure 18, our method demonstrates robust generalization across different training levels. Notably, when the adversarial perturbation is large at a PNR of 0 dB, the proposed method achieves approximately 30$\%$ higher accuracy compared to the AT trained ViT model, whose FNR nearly achieves 100$\%$ under the same conditions. Table \ref{tab:example} compares the computational costs of our proposed AiTViT with several baseline methods for processing 1000 samples, including NT trained ViT, AT trained ViT, NR, and ATARD. The proposed AiTViT achieves competitive performance with a processing time of 0.21 seconds, significantly lower than that of NR (2.43 seconds). Compared with the other four baseline methods, it also demonstrates efficient CPU utilization at 1.4$\%$ and memory usage of 1568 MB, striking a balance between computational efficiency and resource consumption. Notably, AiTViT maintains comparable computational costs to the baseline methods while delivering superior overall performance.
    
\begin{figure}[ht!]
\centering
\includegraphics[width=0.88\columnwidth]{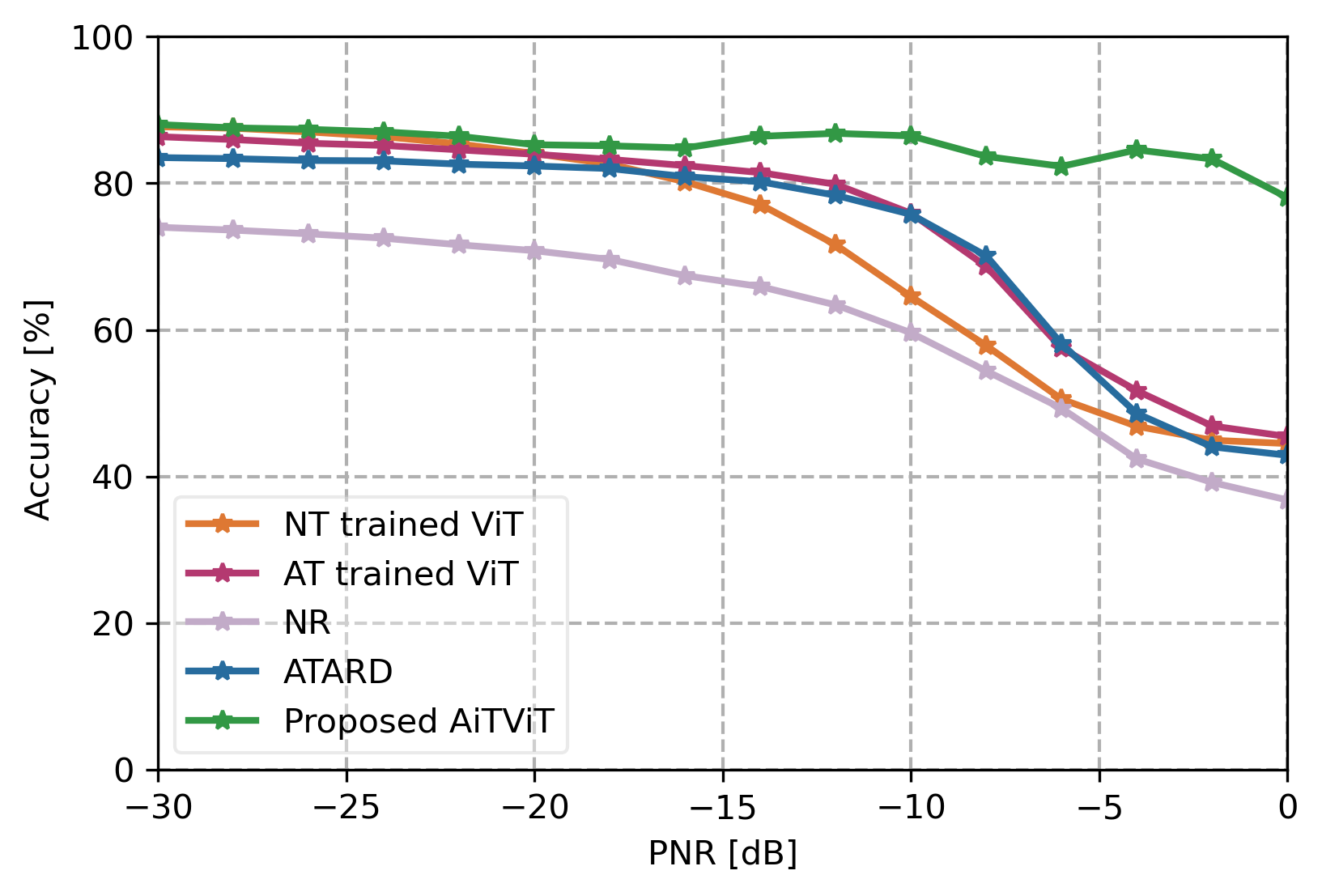}
  \caption{Accuracy against PGD attacks for a range of PNR values.}
  \label{fig:PGD_total}
\end{figure}

\begin{figure}[ht!]
\centering
\includegraphics[width=0.88\columnwidth]{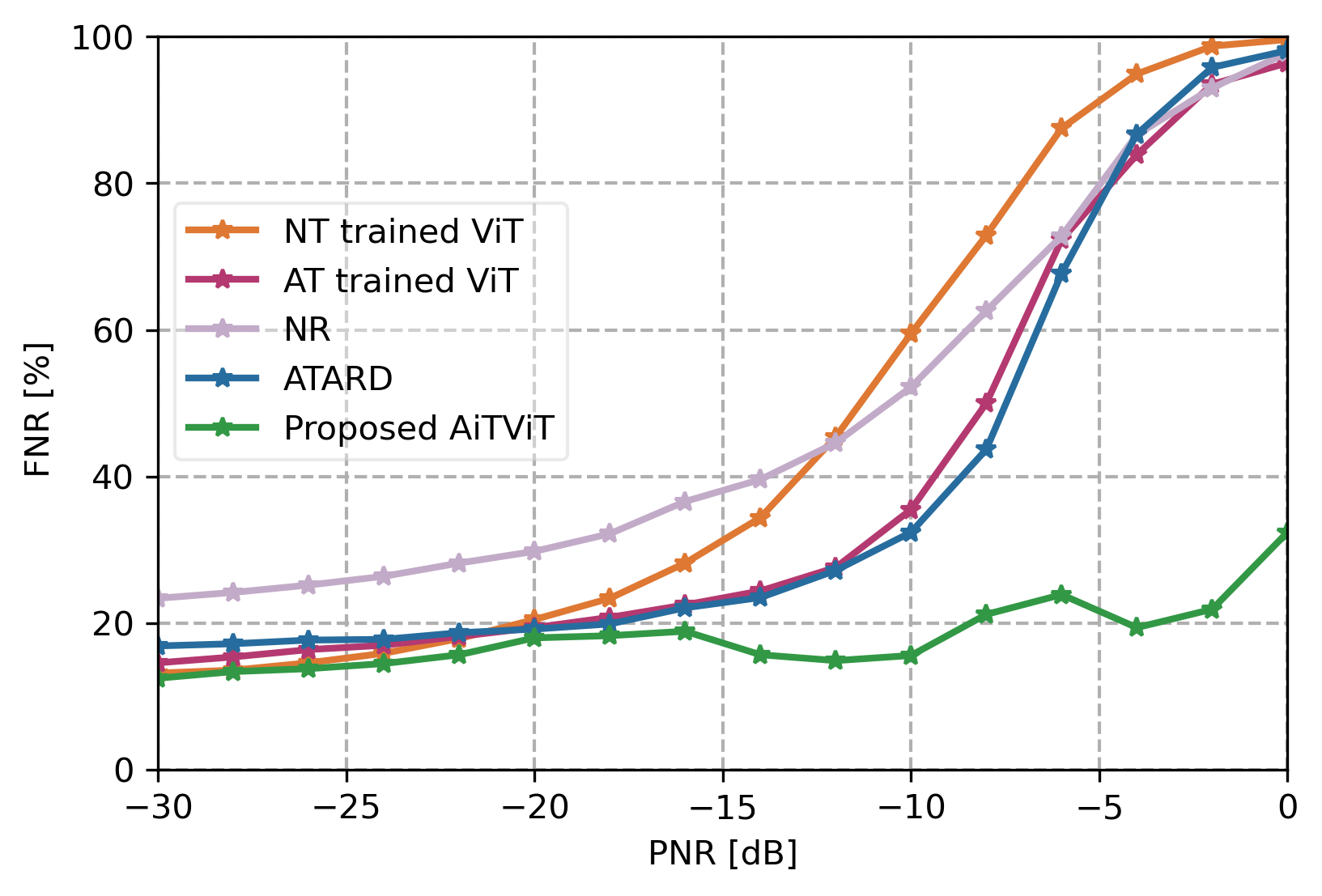}
  \caption{FNR against PGD attacks for a range of PNR values.}
  \label{fig:PGD_total}
\end{figure}

\begin{figure}[ht!]
\centering
\includegraphics[width=0.88\columnwidth]{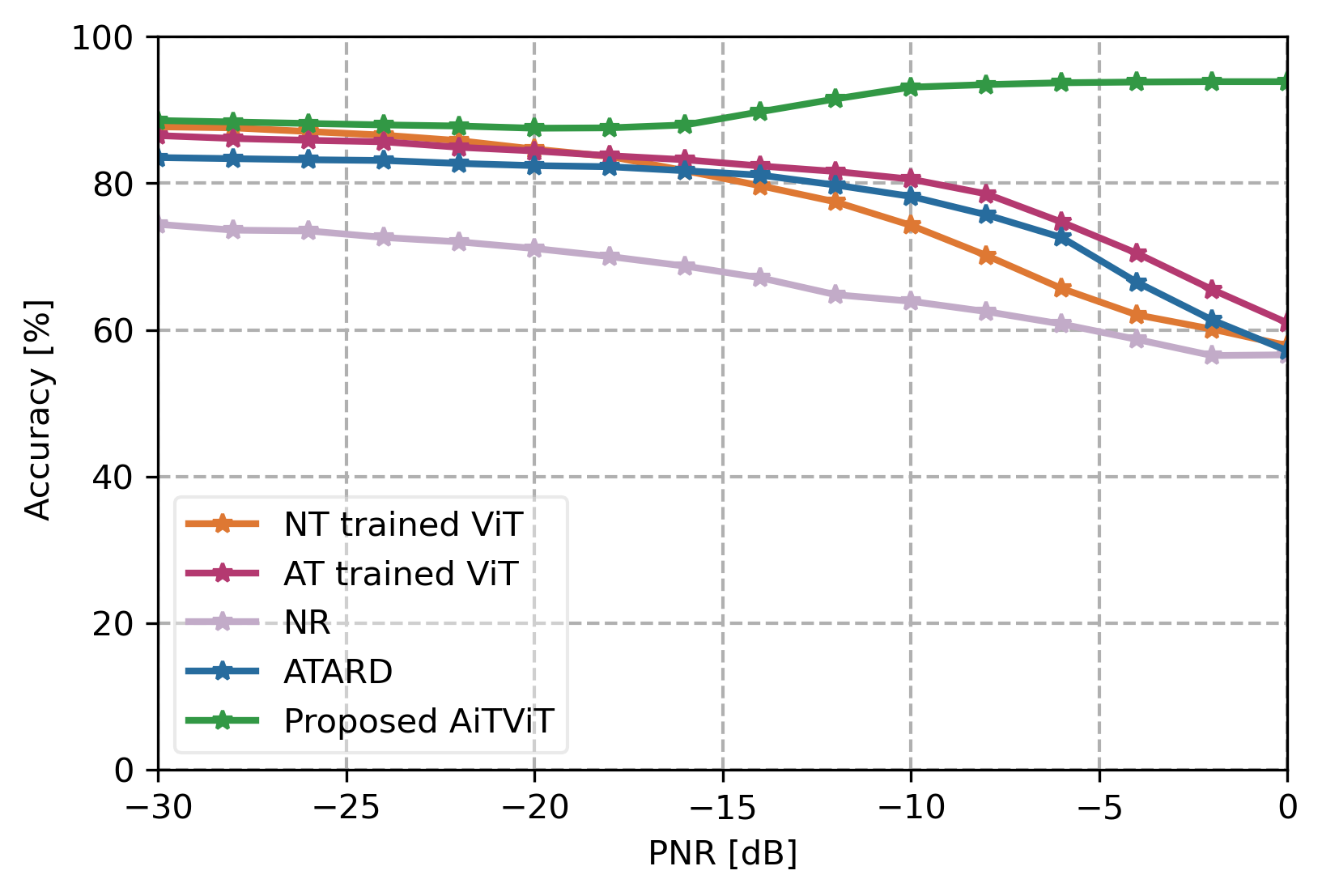}
  \caption{Accuracy against FGM attacks for a range of PNR values.}
  \label{fig:FGM_total}
\end{figure}

\begin{figure}[ht!]
\centering
\includegraphics[width=0.88\columnwidth]{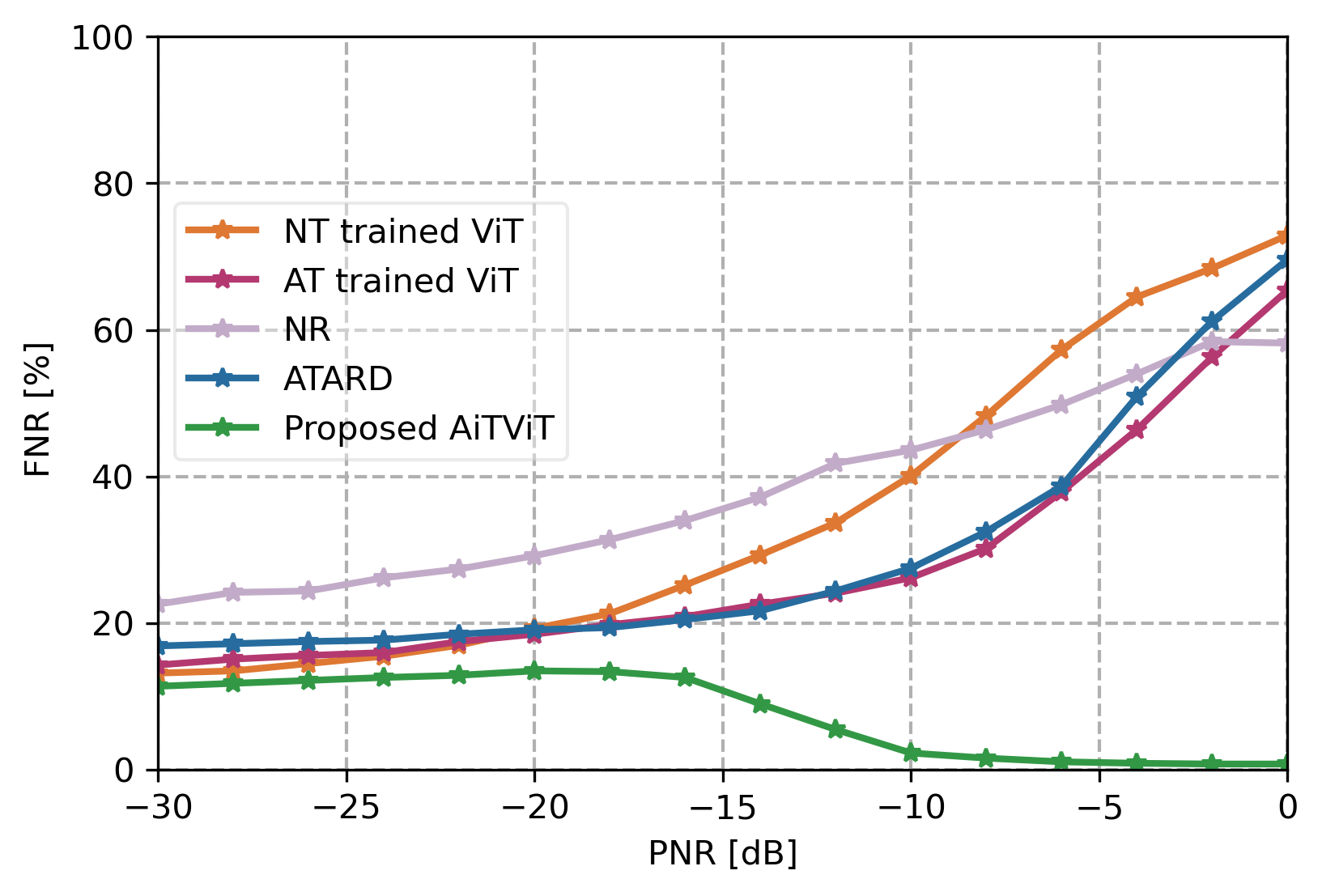}
  \caption{FNR against FGM attacks for a range of PNR values.}
  \label{fig:FGM_total}
\end{figure}

\begin{figure}[ht!]
\centering
\includegraphics[width=0.88\columnwidth]{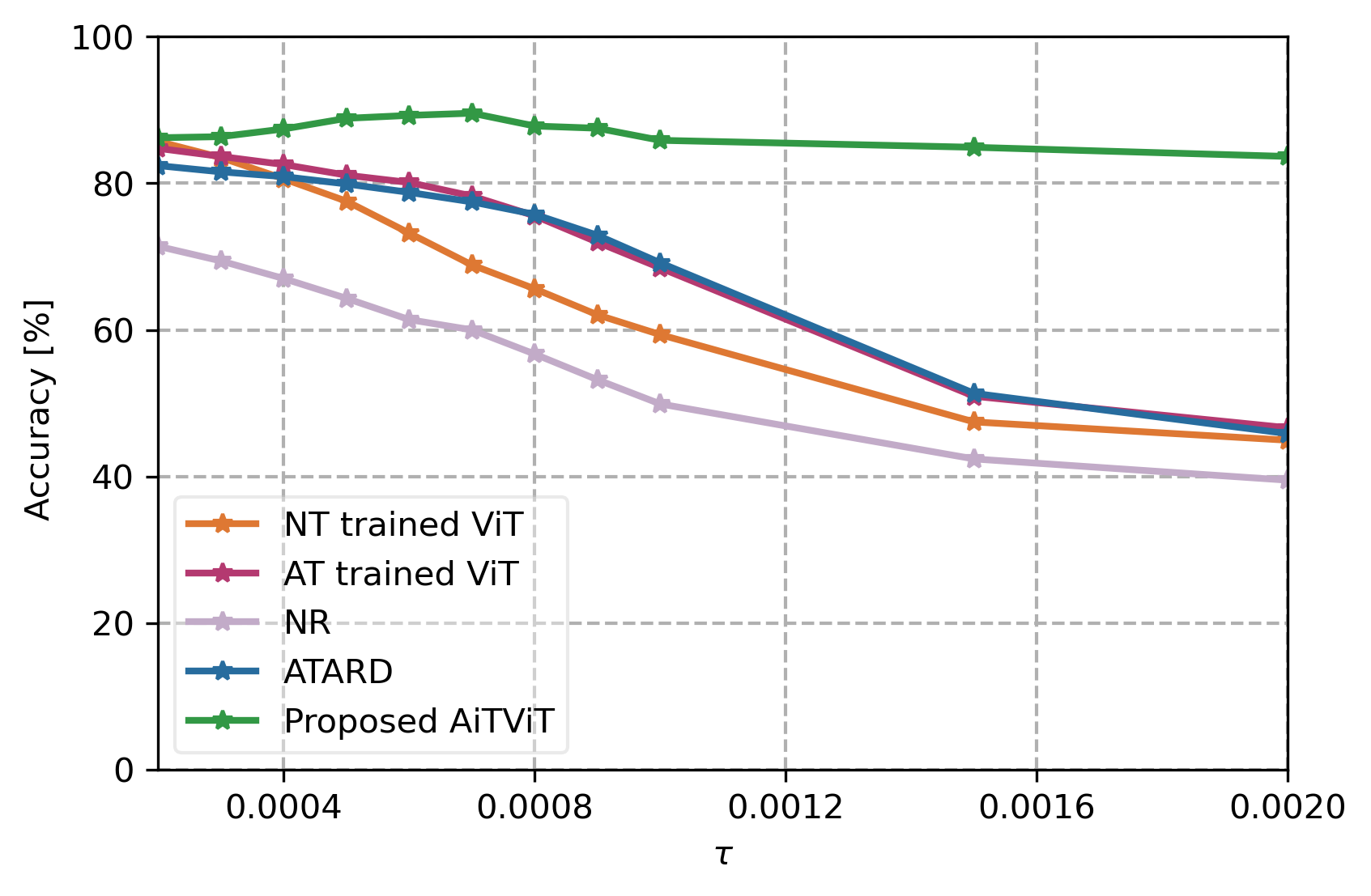}
  \caption{Accuracy against BIM attacks for a range of $\tau$ values.}
  \label{fig:FGM_total}
\end{figure}

\begin{figure}[ht!]
\centering
\includegraphics[width=0.88\columnwidth]{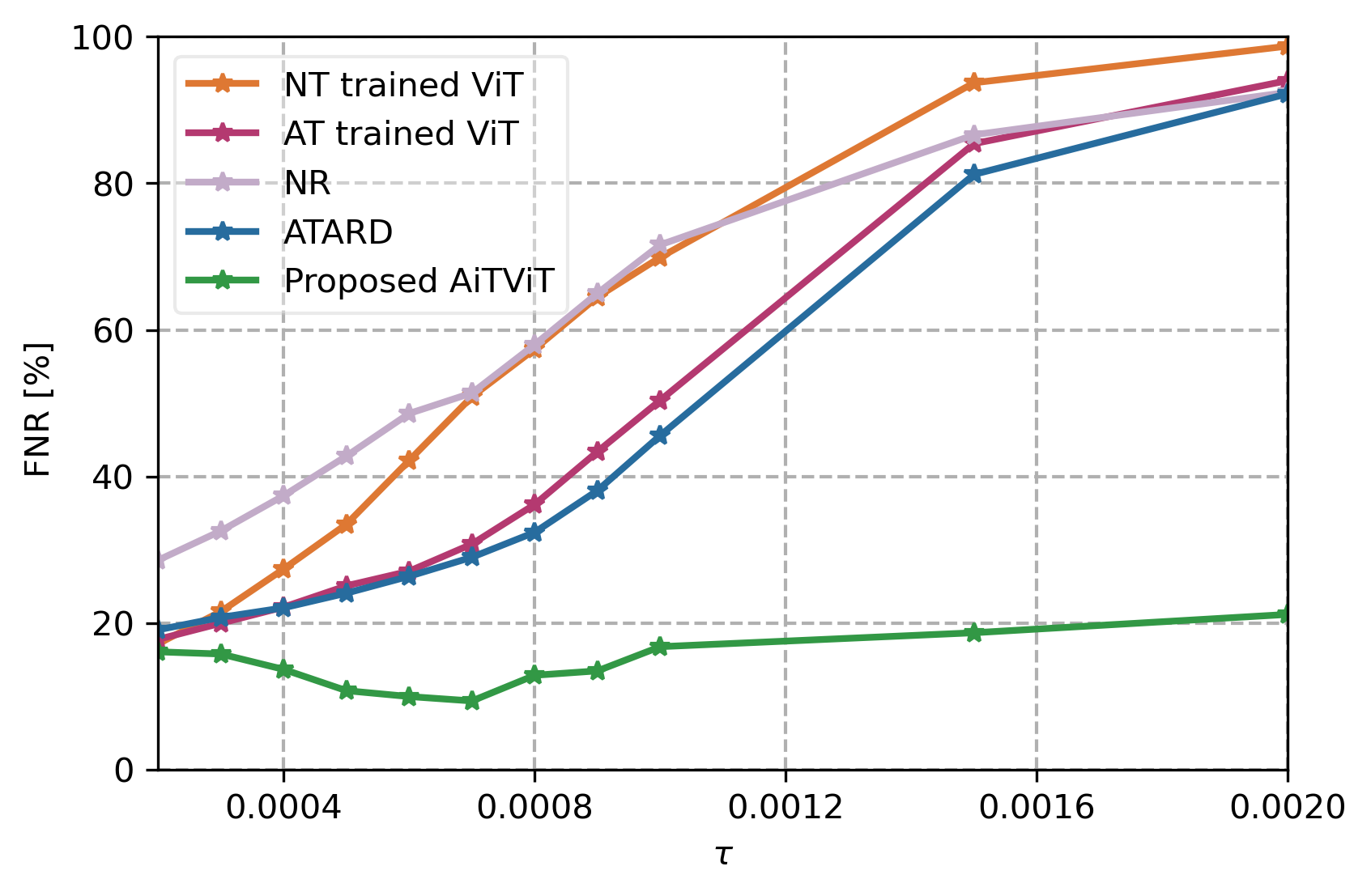}
  \caption{FNR against BIM attacks for a range of $\tau$ values.}
  \label{fig:FGM_total}
\end{figure}

\begin{figure}[ht!]
\centering
\includegraphics[width=0.88\columnwidth]{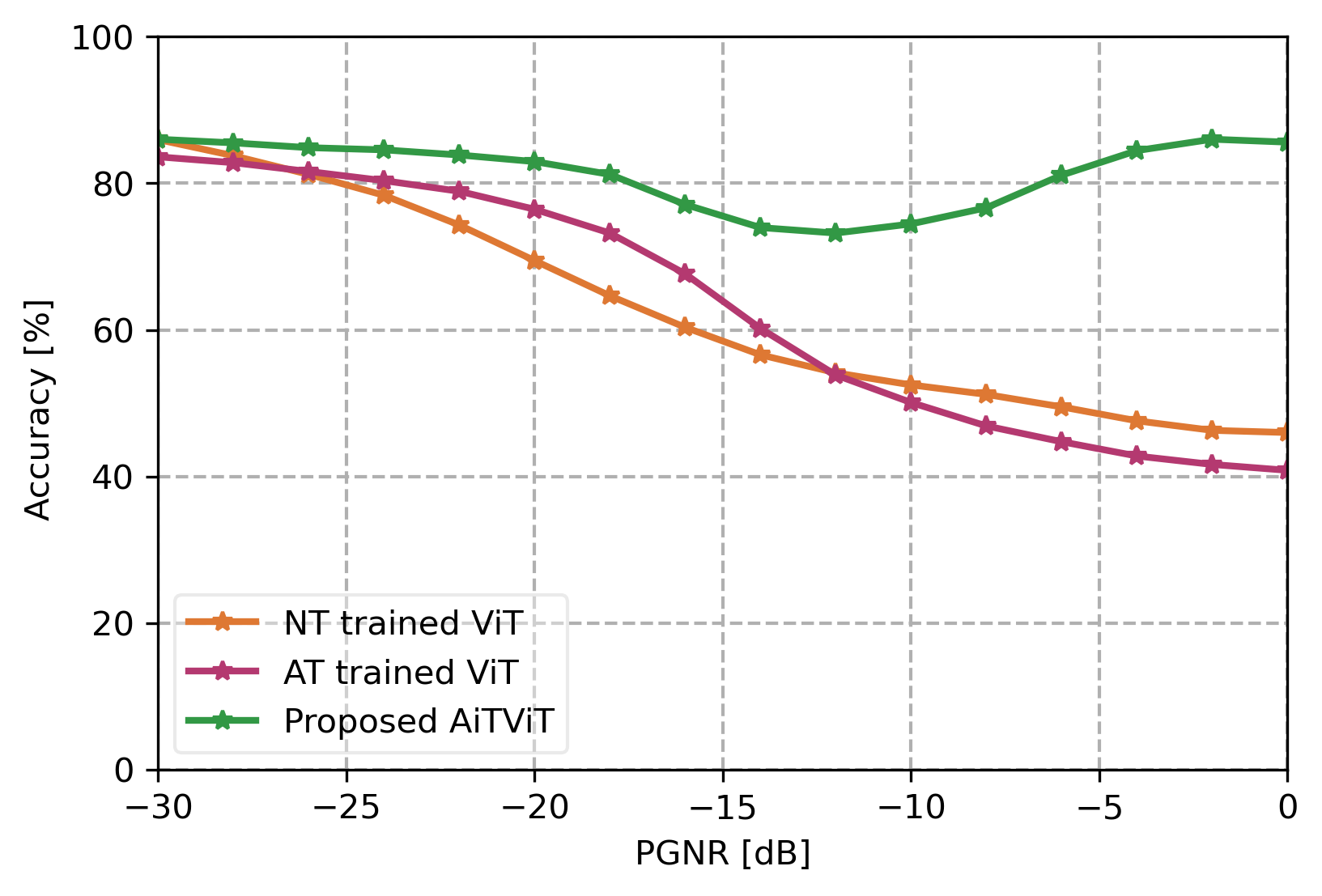}
  \caption{Accuracy against PGD attacks for a range of PGNR values using RDL dataset.}
  \label{fig:PGD_10dB_RDL}
\end{figure}

\begin{figure}[ht!]
\centering
\includegraphics[width=0.88\columnwidth]{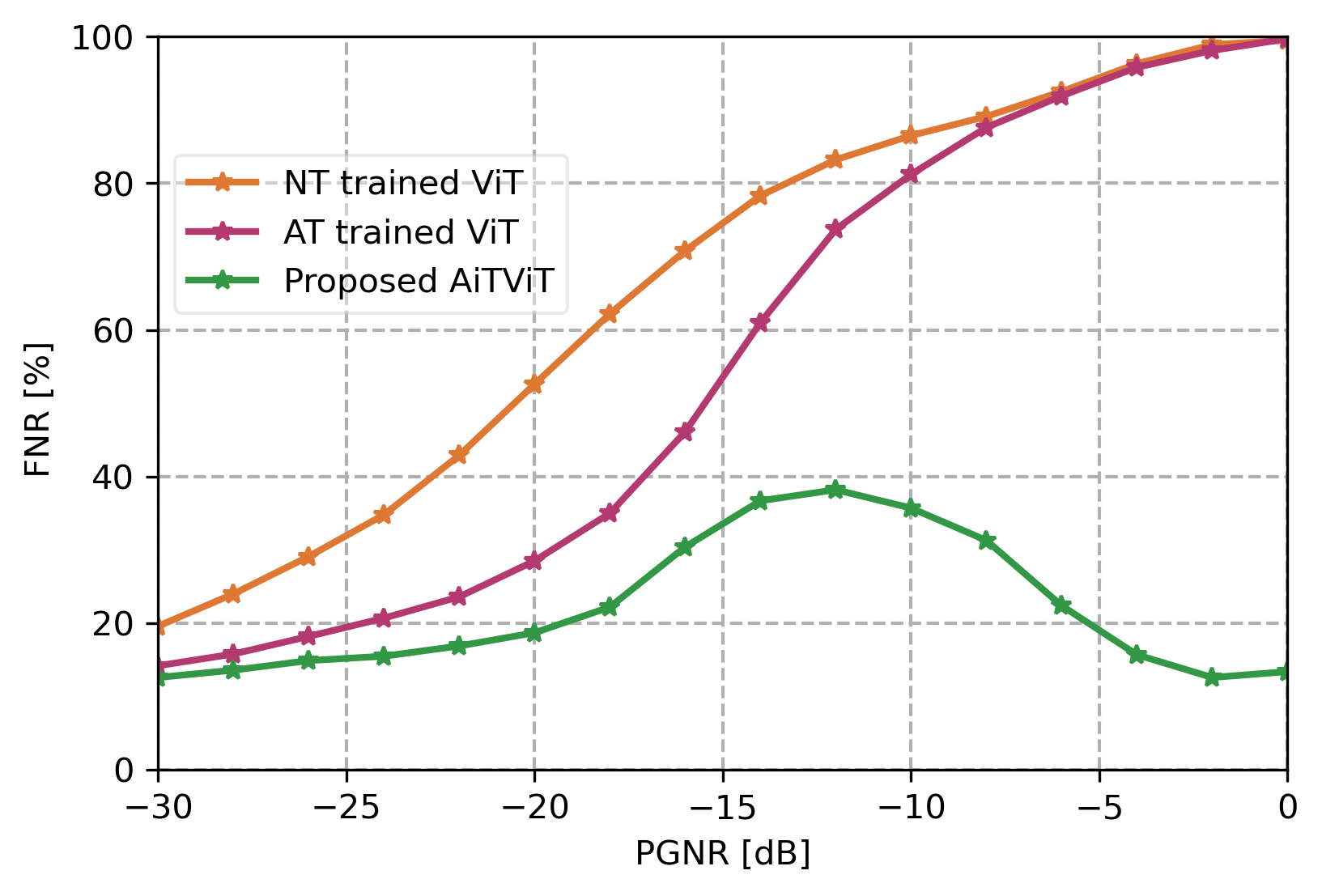}
  \caption{FNR against PGD attacks for a range of PGNR values using RDL dataset.}
  \label{fig:PGD_10dB_RDL_FNR}
\end{figure}

\begin{figure}[ht!]
\centering
\includegraphics[width=0.88\columnwidth]{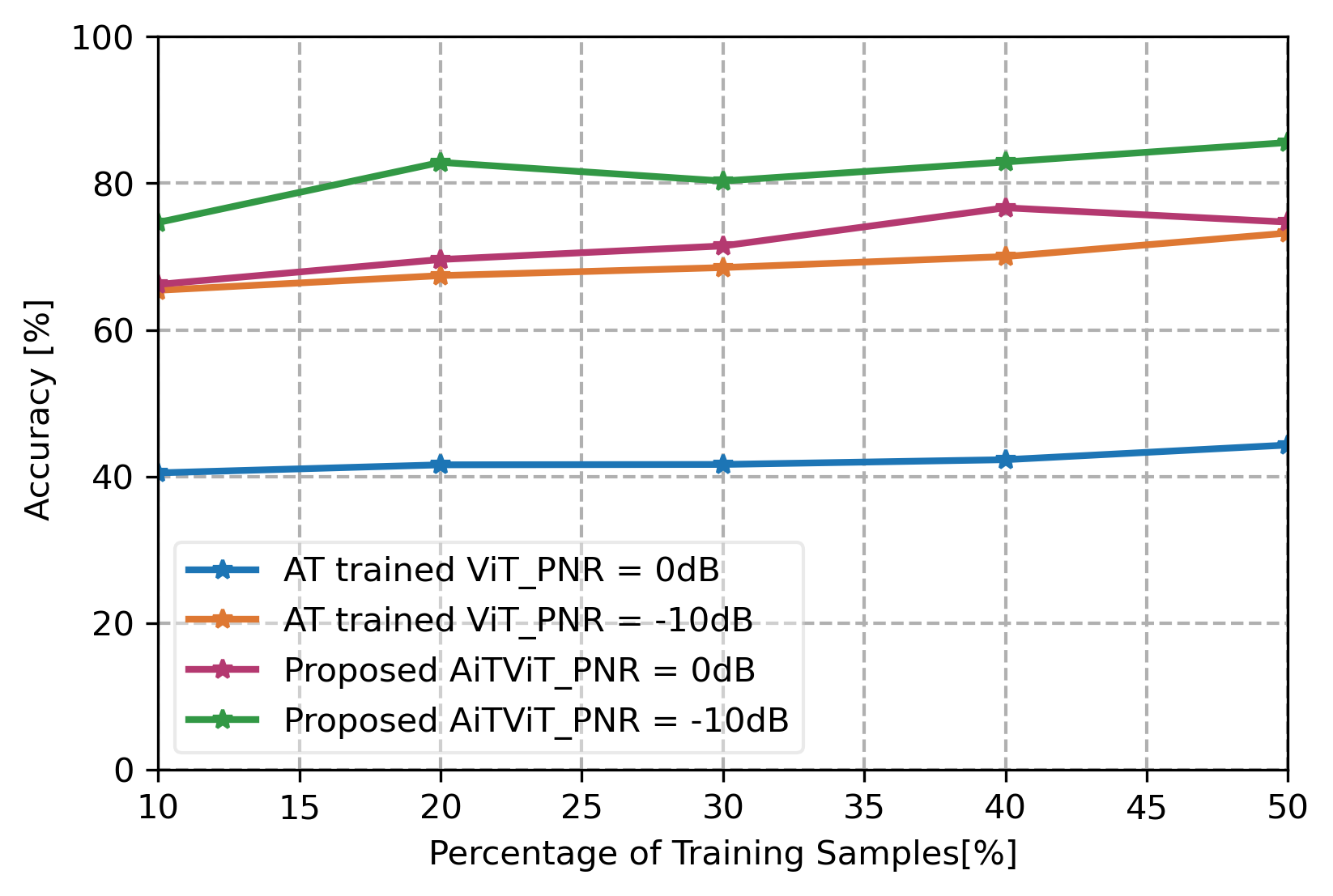}
  \caption{Accuracy of our proposed AiTViT method for a  range of percentage of training samples.}
  \label{fig:generalization}
\end{figure}

\begin{figure}[ht!]
\centering
\includegraphics[width=0.88\columnwidth]{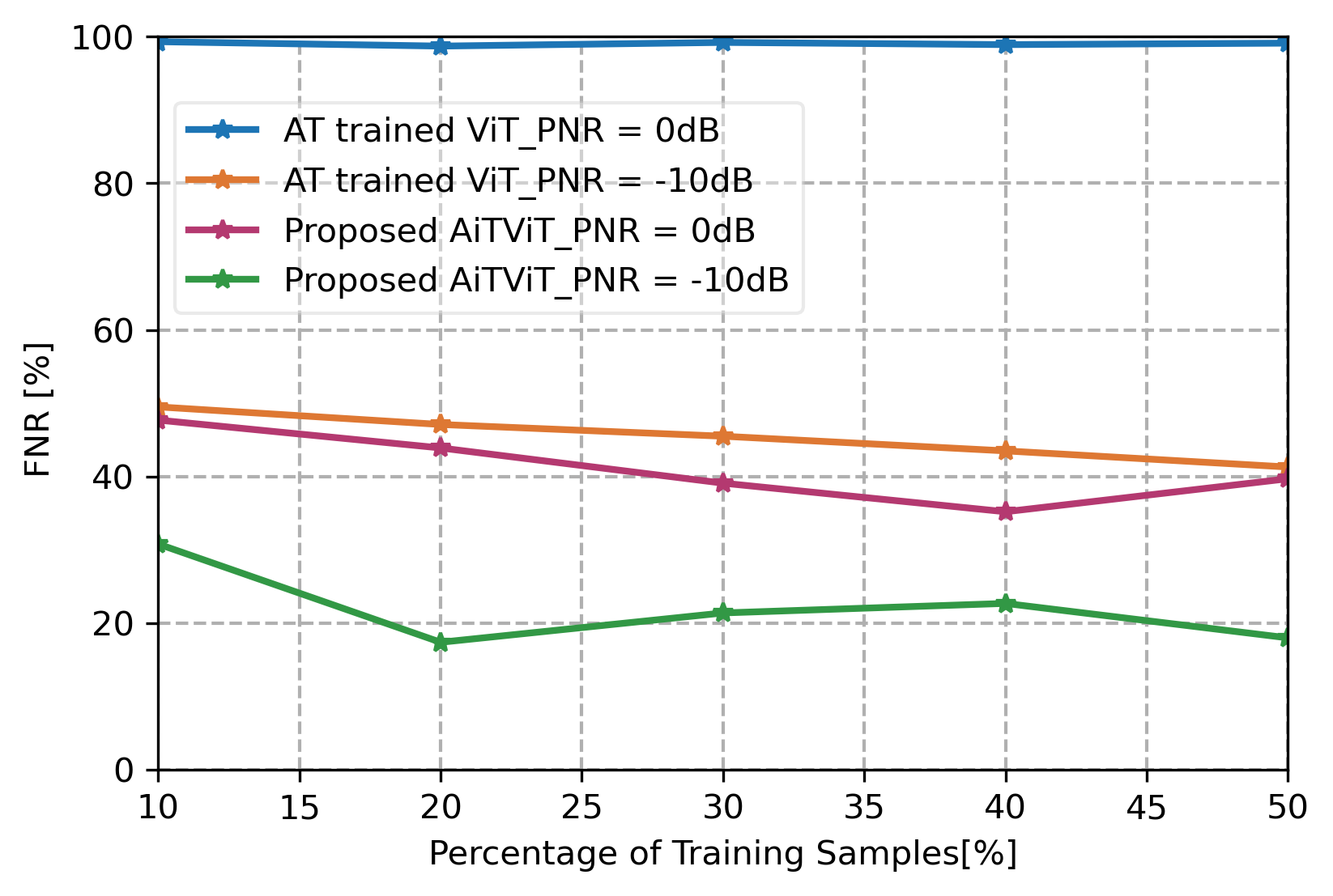}
  \caption{FNR of our proposed AiTViT method for a range of percentage of training samples.}
  \label{fig:generalization}
\end{figure}

\begin{table}[h!]
\centering
\scalebox{0.66}{
\begin{tabular}{|c|c|c|c|}
\hline
         & Processing time (s) & CPU utilization ($\%$) & Memory Usage (MB) \\ \hline
NT trained ViT & 0.21 & 1.4 & 1678 \\ \hline
AT trained ViT & 0.21 & 1.4 & 1678 \\ \hline
NR             & 2.43 & 1.7 & 1339 \\ \hline
ATARD          & 0.09 & 1.2 & 1571 \\ \hline
Proposed AiTViT & 0.21 & 1.4 & 1568 \\ \hline
\end{tabular}}
\caption{Comparison of Computation Costs}
\label{tab:example}
\end{table}

\section{Conclusion}
A novel vision transformer architecture was proposed which includes an adversarial CLS token to detect adversarial attacks. To the best of our knowledge, this is the first work in the literature to propose an adversarial CLS token to defend against adversarial perturbations. The proposed method can integrate a training time defense (i.e., adversarial training method) and a running time defense (i.e., detection mechanism using advI token) into unified network, which reduces the computation complexity as compared to two separate networks as a detection system. We have also examined the impact of our proposed AiTViT on the attention weights, demonstrating how it highlights regions
or features in the input data that may be considered anomalous. Through experimental results, we have shown that the proposed AiTViT achieves a better robustness against adversarial attacks compared to the most competing works in the literature including NT trained ViT, AT trained ViT, NR system and ATARD defense. 

\section*{Acknowledgments}
This work is supported by UKRI through the research grants EP/R007195/1 (Academic Centre of Excellence in Cyber Security Research - University of Warwick) and EP/Y028813/1 (National Hub for Edge AI). S. Lambotharan would like to acknowledge the financial support of the Engineering and Physical Sciences Research Council (EPSRC) projects under grant EP/X012301/1, EP/X04047X/1, and EP/Y037243/1. This work was partially supported by projects SERICS (PE00000014) and FAIR (PE00000013) under the MUR National Recovery and Resilience Plan funded by the European Union - NextGenerationEU.

\small{
    \bibliographystyle{IEEEtran}
    \bibliography{references}
}

\vspace*{-5\baselineskip}

\begin{IEEEbiography}[{\includegraphics[width=1in,height=1.25in,clip,keepaspectratio]{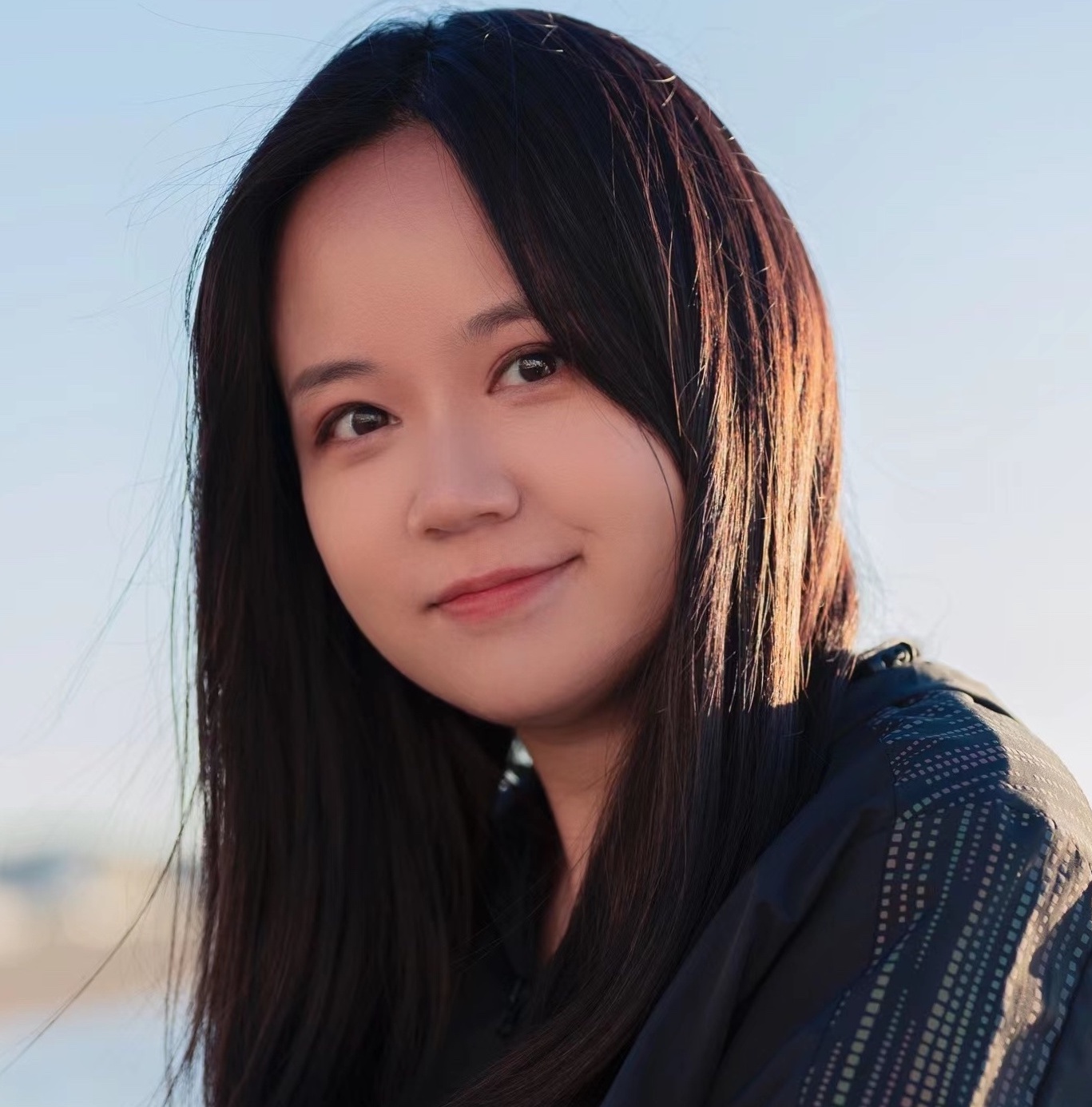}}]{Lu Zhang} is currently a lecturer in the Department of Computer Science, Swansea University. She received B.Eng in Electronic and Information Engineering from Xidian University, China and M.Sc (with distinction) in Mobile Communications from Loughborough University, UK, in 2016 and 2018, respectively. She received Clarke-Griffiths Best Student Prize from Loughborough University in 2018. She received her PhD from Loughborough university in 2022. Prior to this position, she was a research fellow in cyber systems engineering at University of Warwick. Her research interests are wide-ranging, with a primary focus on deep learning and adversarial learning, and their practical applications in areas such as wireless communications and cybersecurity.
\end{IEEEbiography}
\vspace*{-4\baselineskip}

\begin{IEEEbiography}[{\includegraphics[width=1in,height=1.25in,clip,keepaspectratio]{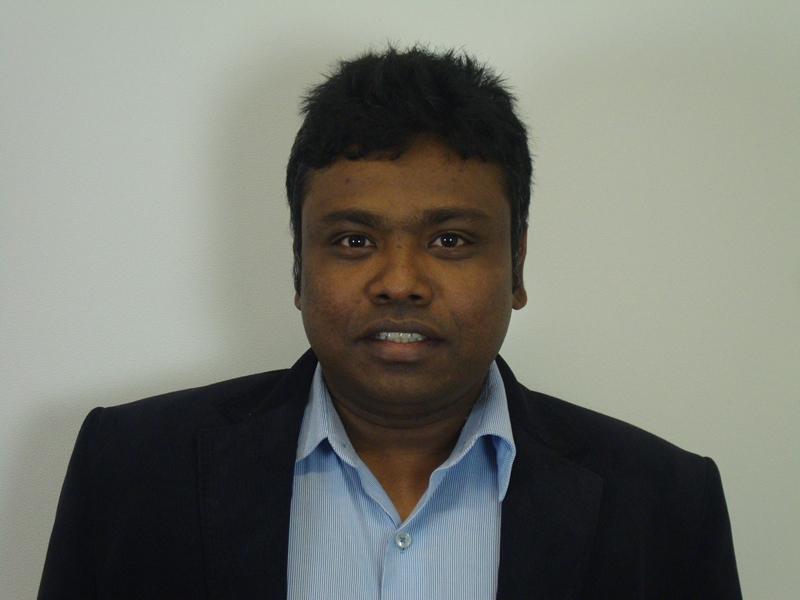}}]{Sangarapillai Lambotharan} is a Professor of Signal Processing and Communications and the Director of the Institute for Digital Technologies at Loughborough University London. He received his Ph.D. in Signal Processing from Imperial College London, U.K., in 1997. In 1996, he was a Visiting Scientist at the Engineering and Theory Centre, Cornell University, USA. Until 1999, he was a Post-Doctoral Research Associate at Imperial College London. From 1999 to 2002, he worked with the Motorola Applied Research Group, U.K., where he investigated various projects, including physical link layer modelling and performance characterization of 2.5G and 3G networks.

He served as a Lecturer at King’s College London and a Senior Lecturer at Cardiff University from 2002 to 2007. His current research interests include 5G networks, MIMO, signal processing, machine learning, and network security. He has authored more than 280 journal articles and conference papers, which have attracted 7200 citations and garnered an h-index of 44. He is a Fellow of the IET and a Senior Member of the IEEE. He currently serves as a Senior Area Editor for IEEE Transactions on Signal Processing and served as an Associate Editor for IEEE Transactions on Communications.
\end{IEEEbiography}
\vspace*{-4\baselineskip}

\begin{IEEEbiography}[{\includegraphics[width=1in,height=1.25in,clip,keepaspectratio]{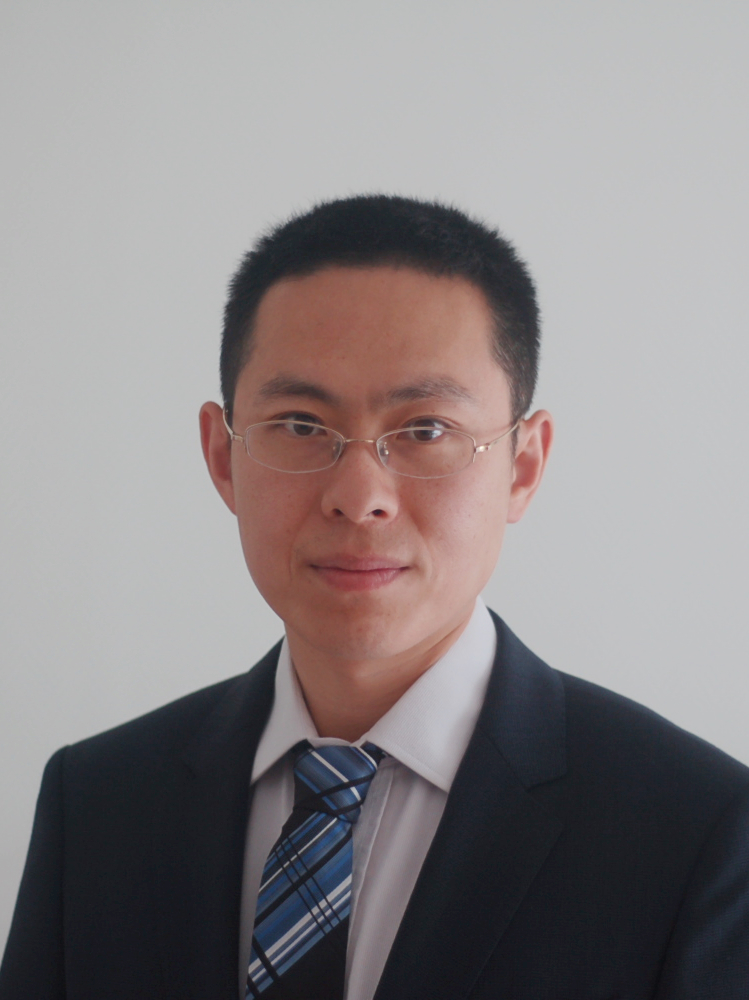}}]{Gan Zheng} (S'05-M'09-SM'12-F'21) received the BEng and the MEng from Tianjin University, Tianjin, China, in 2002 and 2004, respectively, in Electronic and Information Engineering, and the PhD degree in Electrical and Electronic Engineering from The University of Hong Kong in 2008. He is currently Professor in Connected Systems in the School of Engineering, University of Warwick, UK. His research interests include machine learning and quantum computing for wireless communications, reconfigurable intelligent surface, UAV communications and edge computing. He is the first recipient for the 2013 IEEE Signal Processing Letters Best Paper Award, and he also received 2015 GLOBECOM Best Paper Award, and 2018 IEEE Technical Committee on Green Communications \& Computing Best Paper Award. He was listed as a Highly Cited Researcher by Thomson Reuters/Clarivate Analytics in 2019. He currently serves as an Associate Editor for IEEE Wireless Communications Letters and IEEE Transactions on Communications.
\end{IEEEbiography}
\vspace*{-4\baselineskip}

\begin{IEEEbiography}[{\includegraphics[width=1in,height=1.25in,clip,keepaspectratio]{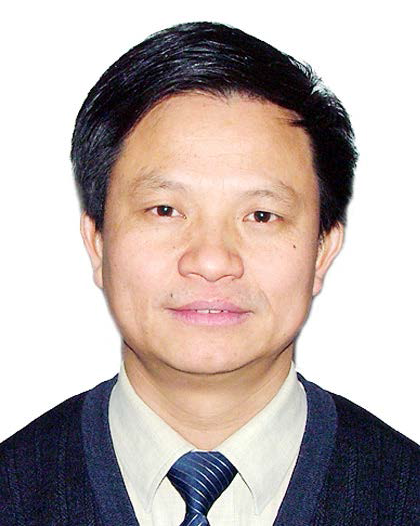}}]{Guisheng Liao} was born in Guangxi, China, in 1963. He received the B.S. degree in mathematics from Guangxi University, Guangxi, China, in 1985, the M.S. degree in computer software from Xidian University, Xi’an, China, in 1990, and the Ph.D. degree in signal and information processing from Xidian University, Xi’an, China, in 1992. From 1999 to 2000, he was a Senior Visiting Scholar with The Chinese University of Hong Kong, Hong Kong. Since 2006, he has been serving as the panelist for the medium- and long-term development plans in high-resolution and remote sensing systems. Since 2007, he has been the Lead of the Chang Jiang Scholars Innovative Team, Xidian University, and devoted to advanced techniques in signal and information processing. Since 2009, he has been the Evaluation Expert for the International Cooperation Project of the Ministry of Science and Technology in China. He was a Chang Jiang Scholars Distinguished Professor with the National Laboratory of Radar Signal Processing and serves as the Dean of the School of Electronic Engineering, Xidian University. He is the author or a coauthor of several books and more than 200 publications. His research interests include array signal processing, space–time adaptive processing, radar waveform design, and airborne/space surveillance and warning radar systems.
\end{IEEEbiography}
\vspace*{-4\baselineskip}

\begin{IEEEbiography}[{\includegraphics[width=1in,height=1.25in,clip,keepaspectratio]{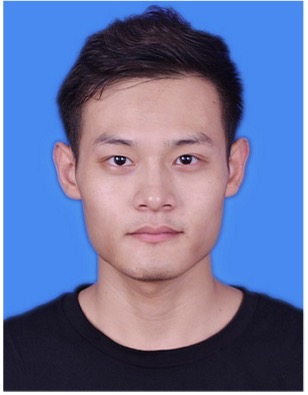}}]{Xuekang Liu} (Member, IEEE) received the M.S. degree (with honors) in Electromagnetic Field and Microwave Technology from Xidian University, Xi’an, China, in 2020, and the Ph.D. degree in Engineering from the University of Kent, Canterbury, U.K., in 2024. He was a Postdoctoral Researcher with the School of Engineering, Lancaster University, from June 2024 to May 2025, and joined Imperial College London as a Postdoctoral Researcher in June 2025.

He was a recipient of the Best Student Paper Award at the 17th International Workshop on Antenna Technology (iWAT 2022), Dublin, and the Outstanding Academic Achievement for his M.S. program. He received the Young Scientist Award from the International Union of Radio Science (URSI) Member Committee, Germany, in 2022. He serves as an active reviewer for more than ten leading international journals and conferences in the field of antennas, microwaves, and wireless systems.
\end{IEEEbiography}
\vspace*{-25\baselineskip}

\begin{IEEEbiography}[{\includegraphics[width=1in,height=1.25in,clip,keepaspectratio]{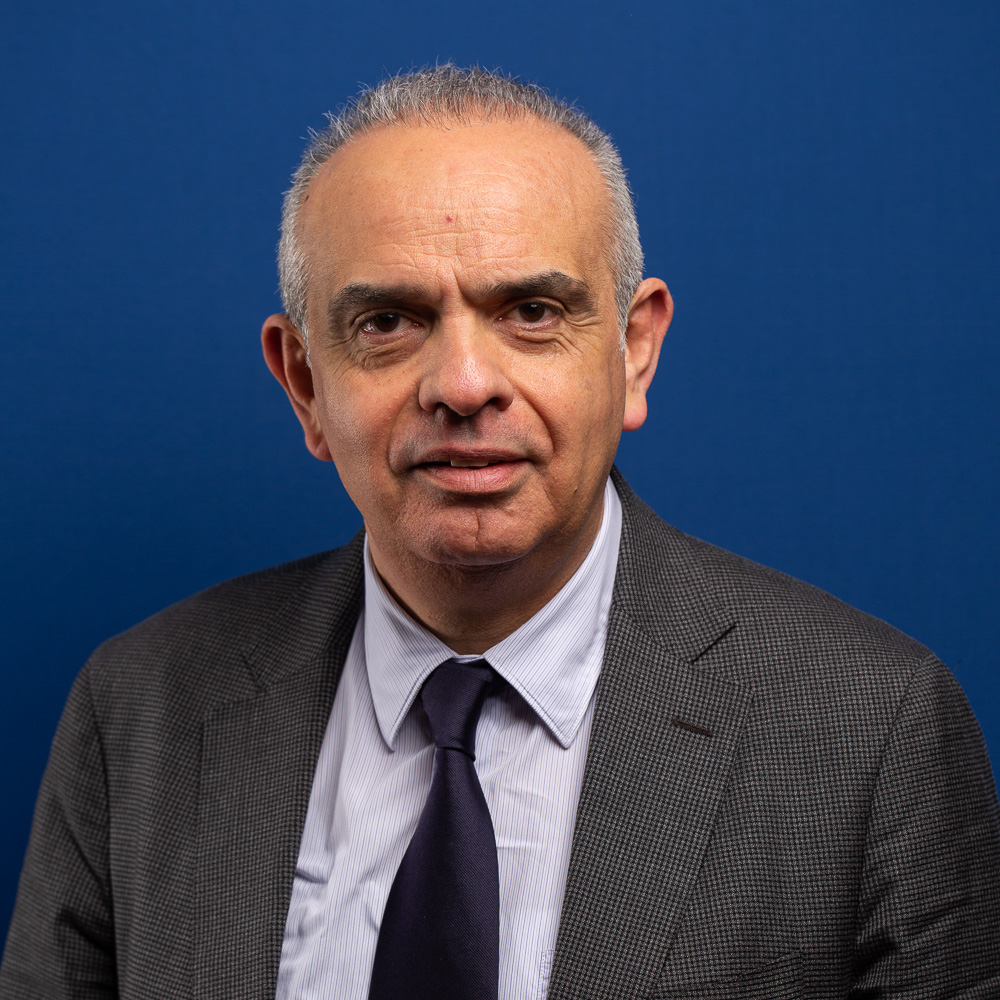}}]{Fabio Roli} is Full Professor of Computer Engineering at the Universities of Genova and Cagliari, Italy. He is Director of the sAIfer Lab, a joint lab between the Universities of Genova and Cagliari on Safety and Security of AI. Fabio Roli’s research over the past thirty years has addressed the design of machine learning systems in the context of real security applications. He has provided seminal contributions to the fields of ensemble learning and adversarial machine learning and he has played a leading role in the establishment and advancement of these research themes. He is a recipient of the Pierre Devijver Award for his contributions to statistical pattern recognition. He has been appointed Fellow of the IEEE, Fellow of the International Association for Pattern Recognition, Fellow of the Asia-Pacific Artificial Intelligence Association.
\end{IEEEbiography}
\vspace*{-25\baselineskip}

\begin{IEEEbiography}[{\includegraphics[width=1in,height=1.25in,clip,keepaspectratio]{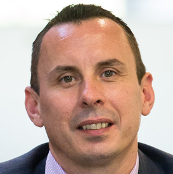}}]{Carsten Maple} is the Director of the NCSC-EPSRC Academic Centre of Excellence in Cyber Security Research and Professor of Cyber Systems Engineering at the University of Warwick. He is also Director for Research Innovation at EDGE-AI, the National Edge Artificial Intelligence Hub, and is a Fellow of the Alan Turing Institute. He has an international research reputation, having published over 450 peer-reviewed papers. He is a member of two Royal Society working groups.
\end{IEEEbiography}


\end{document}